\documentclass[10pt,twocolumn,letterpaper]{article}

\usepackage{wacv/template/wacv}
\usepackage{times}
\usepackage{epsfig}
\usepackage{graphicx}
\usepackage{amsmath}
\usepackage{amssymb}
\usepackage{booktabs}
\usepackage{multirow} 
\usepackage{multicol} 
\usepackage{arydshln} 
\usepackage{float}
\usepackage{enumitem}
\usepackage{color, colortbl}
\usepackage{wrapfig,lipsum,booktabs}
\usepackage{xcolor}

\usepackage{amsmath,amsfonts,bm}

\def\eqref#1{equation~\ref{#1}}

\def\1{\bm{1}}

\DeclareMathAlphabet{\mathsfit}{\encodingdefault}{\sfdefault}{m}{sl}
\SetMathAlphabet{\mathsfit}{bold}{\encodingdefault}{\sfdefault}{bx}{n}

\def\wacvPaperID{582} 

\wacvalgorithmstrack   

\wacvfinalcopy 

\usepackage[pagebackref=true,breaklinks=true,colorlinks,bookmarks=false]{hyperref}

\usepackage[capitalize]{cleveref}
\crefname{section}{Sec.}{Secs.}
\Crefname{section}{Sec.}{Sections}
\Crefname{table}{Table}{Tables}
\crefname{table}{Table}{Tabs.}

\newtheorem{Def sec}{Definition}[section]

\definecolor{mossgreen}{rgb}{0.68, 0.87, 0.68}
\definecolor{teagreen}{rgb}{0.82, 0.94, 0.75}
\definecolor{lightgreen}{rgb}{0.56, 0.93, 0.56}
\definecolor{darkblue}{rgb}{0.0, 0.0, 0.55}
\definecolor{orange-red}{HTML}{de7521}
\definecolor{ForestGreen}{HTML}{00a14b}
\definecolor{lightyellow}{rgb}{1.0, 1.0, 0.88}
\definecolor{upforestgreen}{rgb}{0.0, 0.27, 0.13}
\definecolor{amethyst}{rgb}{0.6, 0.4, 0.8} 
\definecolor{brown(web)}{rgb}{0.65, 0.16, 0.16} 
\definecolor{Gray}{gray}{0.9}
\definecolor{denim}{rgb}{0.08, 0.38, 0.74}
\definecolor{cornflowerblue}{rgb}{0.39, 0.58, 0.93}
\definecolor{powderblue(web)}{rgb}{0.69, 0.88, 0.9}
\definecolor{lightcornflowerblue}{rgb}{0.6, 0.81, 0.93}
\definecolor{columbiablue}{rgb}{0.61, 0.87, 1.0}
\definecolor{lightskyblue}{rgb}{0.53, 0.81, 0.98}

\newcommand{\parahead}[1]{\noindent\textbf{#1.}\enskip}
\newcommand{\quotes}[1]{``#1''}

\newcommand{\edit}[1]{\textcolor{black}{ #1}}
\newcommand{\redit}[1]{\textcolor{black}{ #1}}
\newcommand{\editr}[1]{\textcolor{black}{ #1}}

\begin{document}

\title{Task Agnostic and Post-hoc Unseen Distribution Detection}

\author{
Radhika Dua$^{1}$\qquad Seongjun Yang$^{1}$\qquad Yixuan Li$^{2}$ \qquad Edward Choi$^{1}$\vspace{2mm}\\ 
\text{\normalsize $^1$KAIST\qquad $^2$University of Wisconsin-Madison\qquad}\\
\tt\small\{radhikadua, seongjunyang, edwardchoi\}@kaist.ac.kr \\
\tt\small sharonli@cs.wisc.edu
}

\maketitle
\thispagestyle{empty}

\begin{abstract}
Despite the recent advances in out-of-distribution(OOD) detection, anomaly detection, and uncertainty estimation tasks, there do not exist a task-agnostic and post-hoc approach.
To address this limitation, we design a novel clustering-based ensembling method, called {\bf T}ask {\bf A}gnostic and {\bf P}ost-hoc {\bf U}nseen {\bf D}istribution {\bf D}etection (TAPUDD) that utilizes the features extracted from the model trained on a specific task. Explicitly, it comprises of \emph{TAP-Mahalanobis}, which clusters the training datasets' features and determines the minimum Mahalanobis distance of the test sample from all clusters. Further, we propose the \emph{Ensembling module} that aggregates the computation of iterative \emph{TAP-Mahalanobis} for a different number of clusters to provide reliable and efficient cluster computation.
Through extensive experiments on synthetic and real-world datasets, we observe that our \edit{task-agnostic} approach can detect unseen samples effectively across diverse tasks and performs better or on-par with the existing \edit{task-specific} baselines. 
We also demonstrate that our method is more viable \edit{even} for large-scale classification tasks.
\end{abstract}

\vspace{-2mm}
\section{Introduction}

Deep neural networks have achieved phenomenal performance in diverse domains such as computer vision and healthcare~\cite{bojarski2016end,mikolov2013efficient,dermato16deep}. However, they struggle to handle samples from an unseen distribution, leading to unreliable predictions and fatal errors in safety-critical applications.
In an ideal situation, a robust model should be capable of making predictions on samples from the learned distributions, and at the same time, flag unknown inputs from unfamiliar distributions so that humans can make a responsible decision. 
For instance, in safety-critical tasks such
as cancer detection, the machine learning assistant must issue a warning and hand over the control to the doctors when it detects an unusual sample that it has never seen during training. Thus, in practice, it is important for a  model to know when \textit{not} to predict. This task of detecting samples from an unseen distribution is referred to as out-of-distribution (OOD) detection~\cite{hendrycks2017base,liang2017enhancing,mahala18,sastry2021gram,liu2020energy,Huang2021MOSTS,Hendrycks2019ABF,sehwag2021ssd,Xiao2021DoWR,mahmood2020multiscale}.

Most of these OOD detection methods mainly focusing on classification tasks have shown great success. However, they are not directly applicable to other tasks like regression.
Although a few bayesian and non-bayesian techniques ~\cite{gal2016dropout,lakshminarayanan2017simple,maddox_2019_simple,Graves2011PracticalVI} estimate uncertainty in regression tasks, they are not post-hoc as it often requires a modification to the training pipeline, or multiple trained copies of the model, or training a model with an optimal dropout rate.
This raises an under-explored question:
\begin{quote}
\vspace{-1.2mm}
\textit{Can we design a task-agnostic, and
post-hoc approach for unseen distribution detection 
?}
\vspace{-1.2mm}
\end{quote}

Motivated by this, we propose a novel clustering-based ensembling framework, \quotes{{\bf T}ask {\bf A}gnostic and {\bf P}ost-hoc {\bf U}nseen {\bf D}istribution {\bf D}etection (TAPUDD)}, which comprises of two modules, \emph{TAP-Mahalanobis} and \emph{Ensembling}. \emph{TAP-Mahalanobis} partitions the training datasets' features into clusters and then determines the minimum Mahalanobis distance of a test sample from all the clusters. 
The \emph{Ensembling }module aggregates the outputs obtained from \emph{TAP-Mahalanobis} iteratively for a different number of clusters. It enhances reliability and eliminates the need to determine an optimal number of clusters. \edit{
Since TAPUDD is a post-hoc approach and doesn't require training the model, it is more efficient and easy to deploy in real-world.
}

To demonstrate the efficacy of our approach, we conduct experiments on 2-D synthetic datasets for binary and multi-class classification tasks and observe that our method effectively detects the outliers in both tasks. Further, we extensively evaluate our approach on real-world datasets for diverse tasks. In particular, we evaluate our approach for binary classification (gender prediction) and regression (age prediction) task on RSNA boneage dataset to detect the samples shifted by brightness. We observe that our method successfully identifies the shifted samples. We also evaluate our approach on large-scale classification tasks and obtained logical performance on diverse OOD datasets. To sum up, our contributions include:
\begin{itemize}
\setlength{\itemsep}{1.5pt}
\setlength{\parskip}{0pt}
\setlength{\parsep}{0pt}

\item We propose a novel task-agnostic and post-hoc approach, \textbf{TAPUDD}, to detect unseen samples across diverse tasks like classification, regression, etc. 
\item For the first time, we empirically show that a single approach can be used for multiple tasks with stable performance. 
We conduct exhaustive experiments on synthetic and real-world datasets for regression, binary classification, and multi-class classification tasks to demonstrate the effectiveness of our method.
\item \edit{We conduct ablation studies to illustrate the effect of number of clusters in \emph{TAP-Mahalanobis} and ensembling strategies in TAPUDD. We observe that TAPUDD
performs better or on-par with \emph{TAP-Mahalanobis} 
and eliminates the necessity to determine the optimal number of clusters.}
\end{itemize} 
\section{Related Work}
\label{sec:related_work}
To enhance the reliability of deep neural networks, there exist several efforts along the following research directions:

\parahead{Out-of-distribution Detection}
Recent works have introduced reconstruction-error based~\cite{Schlegl2017UnsupervisedAD,Zong2018DeepAG,Deecke2018ImageAD,Pidhorskyi2018GenerativePN,Perera2019OCGANON,Choi2020NoveltyDV}, density-based~\cite{ren2019likelihood,choi2018waic,mahmood2020multiscale,Du2019ImplicitGA,Grathwohl2020YourCI,Nalisnick2019DetectingOI,Serr2020InputCA}, and self-supervised~\cite{Golan2018DeepAD,Hendrycks2019UsingSL,Bergman2020ClassificationBasedAD,sehwag2021ssd} OOD detection methods. Other efforts
include post-hoc methods~\cite{mahala18,liu2020energy,hendrycks2017base,liang2017enhancing,sastry2021gram,Hendrycks2019ABF,Nalisnick2019DoDG} that do not require modification to the training procedure. 
However, there is no approach that is post-hoc and does not require the class label information of the training data.

\parahead{Uncertainty Estimation} 
Research in this direction primarily estimates the uncertainty to enhance the robustness of networks in regression tasks.
Well-known methods to estimate uncertainty include bayesian~\cite{Mackay1992BayesianMF,Neal1995BayesianLF,Blundell2015WeightUI,Graves2011PracticalVI,Louizos2016StructuredAE,Hasenclever2017DistributedBL,Li2015StochasticEP,korattikara2015bayesian,Welling2011BayesianLV,Springenberg2016BayesianOW,maddox_2019_simple} and non-bayesian~\cite{gal2016dropout,lakshminarayanan2017simple} approaches, which have shown remarkable success.
However, they require significant modification to the training pipeline, multiple trained copies of the model, and are not post-hoc.

\parahead{Anomaly Detection} 
This task aims to detect anomalous samples shifted from the defined normality. Prior work~\cite{Perera2019OCGANON,Wang2019EffectiveEU,Golan2018DeepAD,Schlegl2017UnsupervisedAD,Chalapathy2018AnomalyDU,Zong2018DeepAG,Deecke2018ImageAD} proposed methods
to solve anomaly detection.
However, more recently, ~\cite{sehwag2021ssd,Tack2020CSIND,Bergman2020ClassificationBasedAD,HernndezLobato2015ProbabilisticBF} proposed a unified method to solve both OOD detection and anomaly detection. Nonetheless, these methods require end-to-end training and are not post-hoc.

There exist no unified approach to enhance the reliability of neural networks across distinct tasks like classification, regression, etc.
In contrast to all the aforementioned efforts, our work presents a post-hoc, and task-agnostic approach to detect unknown samples across varied tasks.
\section{Background and Methodology}
\label{sec:background_method}

\subsection{Problem Formulation} 
\label{sec:problem_formulation}
We assume that the in-distribution data $\mathcal{D}_{\text{IN}} =\{ X, Y \}$ is composed of $N$ i.i.d. data points with inputs $X=\{\mathbf{x}_1, ..., \mathbf{x}_N\}$ and labels $Y=\{y_1, ..., y_N\}$.
Specifically, $\mathbf{x}_i\in \mathbb{R}^{d}$ represents a $d$-dimensional input vector, and $y_i$ is its corresponding label. For classification problems, the label $y_i$ is one of the $C$ classes.
For regression problems, the label is real-valued, that is $y_i\in \mathbb{R}$. For autoencoding (or self-supervised or unsupervised learning tasks), the label $y_i$ is absent.
Let $f : X \rightarrow \mathcal{Z}$, where $\mathcal{Z}\in \mathbb{R}^{m}$,  denote the feature extractor often parameterized by a deep neural network which maps a sample
from the $d$-dimensional input space \edit{($X$)} to the $m$-dimensional feature space ($\mathcal{Z}$). 

Our goal is to obtain the feature of a sample from a trained DNN
using the feature extractor ($f$), and equip it with an OOD detector which can detect samples from different distribution than the training distribution (OOD samples) or samples shifted from the training distribution based on some attribute
\cite{Park2021NaturalAS}. We wish to design a task-agnostic, and post-hoc OOD detector which only requires the features of training data obtained from the trained model. Since such OOD detector does not require the raw training data, it minimizes privacy risk, a significant advantage for tasks where data-related privacy is a serious concern.

\subsection{Background}

\parahead{Mahalanobis distance-based OOD detection method} Mahalanobis OOD detector~\cite{mahala18} approximates each class as multi-variate gaussian distribution and use the minimum Mahalanobis distance of a test sample from all classes for OOD detection. \editr{Given that density estimation in high-dimensional space is a known intractable problem, Mahalanobis approach is viewed as a reasonable approximation that has been widely adopted.}
In particular, it fits the gaussian distribution to the features of the  training  data $X=\{(\mathbf{x}_1, y_1), \ldots, (\mathbf{x}_N, y_N)\}$ and compute per-class mean 
\resizebox{.4\hsize}{!}{ $\mathcal{\mu}_{c} = \frac{1}{N_{c}} \sum_{i:y_{i}=c}^{} f(\mathbf{x}_{i})$ } and a shared covariance matrix 
\resizebox{.8\hsize}{!}{ $\Sigma = \frac{1}{N} \sum_{c=1}^{C} \sum_{i:y_{i}=c}^{} (f(\mathbf{x}_{i}) - \mathcal{\mu}_{c}) (f(\mathbf{x}_{i}) - \mathcal{\mu}_{c})^{T},$ }
where $f(\mathbf{x}_{i})$ 
denotes the penultimate layer features of an input sample $\mathbf{x}_i$, $C$ denotes the total number of classes in the training dataset, and $N_c$ denotes the total number of samples with class label $c$.
Given a test sample $\mathbf{x}_{test}$, the mahalanobis score is defined as:
\vspace{-0.8mm}
\begin{equation}
\resizebox{0.95\hsize}{!}{
$\mathcal{S}_{\text{Mahalanobis}} = -\min\limits_{c} (f(\mathbf{x}_{test}) - \mathcal{\mu}_{c})^{T} \Sigma ^{-1}(f(\mathbf{x}_{test}) - \mathcal{\mu}_{c}),$
\nonumber
}
\vspace{-0.8mm}
\end{equation}
where $f(\mathbf{x}_{test})$ denotes the penultimate layer features of a test sample $\mathbf{x}_{test}$.

\begin{figure*}[t]
\centering
    \vspace{-2mm}
     \includegraphics[width=0.9\textwidth]{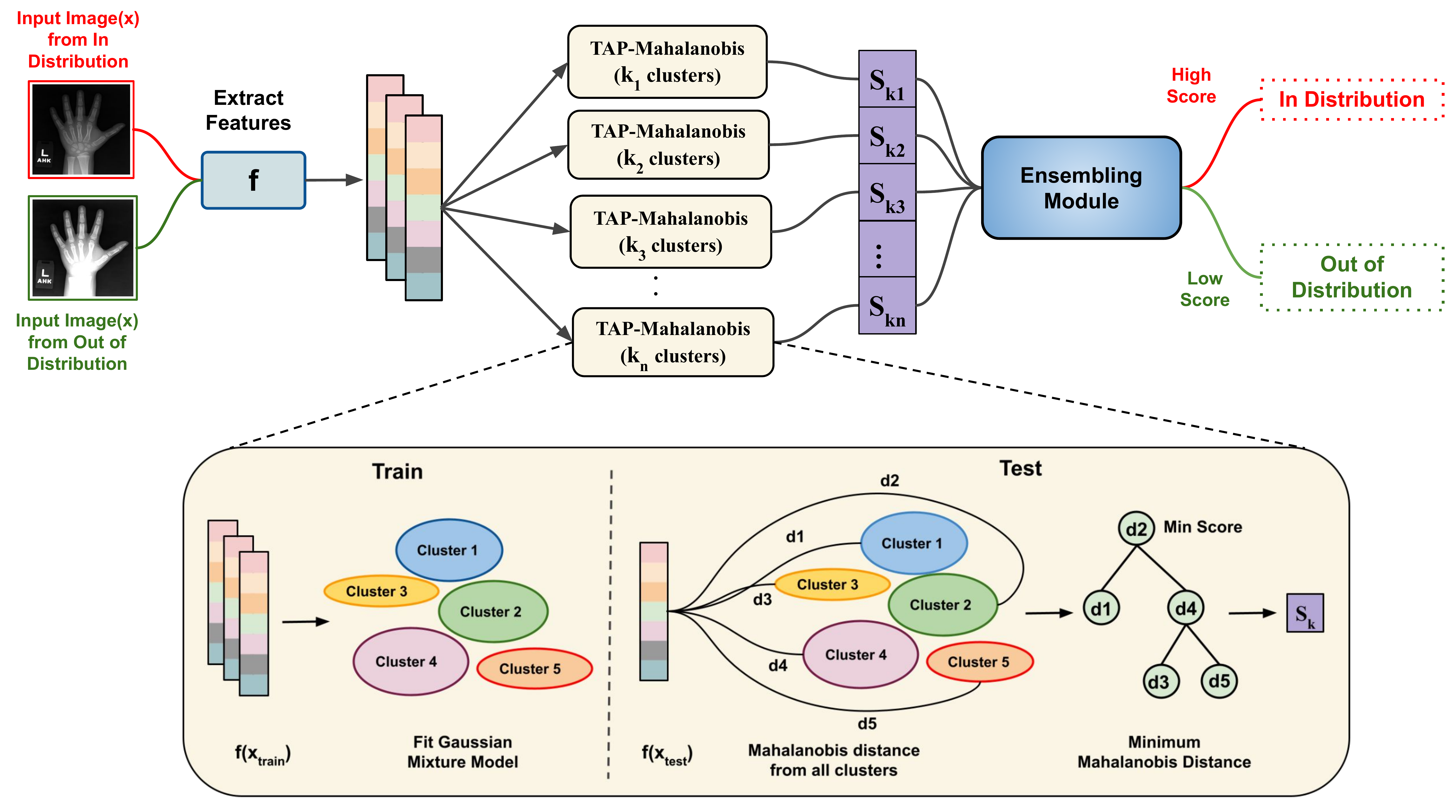}
     \vspace{-1mm}
     \caption{
    \edit{\textbf{TAPUDD.} Our method first extracts the features of an input image $\mathbf{x}$ from the feature extractor $f$ of a model trained on a specific task. \emph{TAP-Mahalanobis} module then uses the extracted features $f(\mathbf{x}_{train})$ to fit the gaussian mixture model and computes the minimum mahalanobis distance $S_k$ for the given feature vector $f(\mathbf{x}_{test})$.
    Further, the \emph{Ensembling} module aggregates the mahalanobis distance ($S_{k1}$ to $S_{kn}$) obtained from iterative computation of \emph{TAP-Mahalanobis} for different number of clusters ($k_1$ to $k_n$)
    to enhance the reliability.
    } 
     }
     \label{fig:method}
\end{figure*}

\subsection{TAPUDD: {\bf T}ask {\bf A}gnostic and {\bf P}ost-hoc {\bf U}nseen {\bf D}istribution {\bf D}etection}
\label{sec:tapudd}

We propose a novel, {\bf T}ask {\bf A}gnostic and {\bf P}ost-hoc {\bf U}nseen {\bf D}istribution {\bf D}etection (TAPUDD)  method, as shown in \cref{fig:method}. The method comprises of two main modules \emph{TAP-Mahalanobis} and \emph{Ensembling}.\\

\parahead{TAP-Mahalanobis} Given training samples $X=\{\mathbf{x}_1, ..., \mathbf{x}_N\}$, we extract the features of the in-distribution data from a model trained for a specific task using a feature extractor $f$. We then pass these features to the \emph{TAP-Mahalanobis} module. It first partition the features of the in-distribution data into $K$ clusters using Gaussian Mixture Model (GMM) with \quotes{\textit{full}} covariance. Then, we model the  features in each cluster independently as multivariate gaussian and compute the empirical cluster mean and covariance of training samples $X=\{\mathbf{x}_1, ..., \mathbf{x}_N\}$ 
and their corresponding cluster labels $C=\{c_1, ..., c_N\}$ as:
\vspace{-0.8mm}
\begin{equation}
\resizebox{.98\hsize}{!}{
    $\mathcal{\mu}_{c} = \frac{1}{N_{c}} \sum_{i:c_{i}=c}^{}f(\mathbf{x}_{i}),
    \Sigma_{c} = \frac{1}{N_{c}} \sum_{i:c_{i}=c}^{} (f(\mathbf{x}_{i}) - \mathcal{\mu}_{c}) (f(\mathbf{x}_{i}) - \mathcal{\mu}_{c})^{T},$
    \nonumber
    }
\vspace{-0.8mm}
\end{equation}
where $f(\mathbf{x}_{i})$ denotes the penultimate layer features of an input sample $\mathbf{x}_i$ from a cluster $c_i$.

Then, given a test sample, $\mathbf{x}_{test}$, we obtain the negative of the minimum of the Mahalanobis distance from the center of the clusters as follows:
\vspace{-0.8mm}
\begin{equation}
\resizebox{.97\hsize}{!}{
    $\mathcal{S}_{\text{TAP-Mahalanobis}} = -\min \limits_{c} (f(\mathbf{x}_{test}) - \mathcal{\mu}_{c})^{T} \Sigma_{c} ^{-1} (f(\mathbf{x}_{test}) - \mathcal{\mu}_{c}),$
    \nonumber
    }
\vspace{-0.8mm}
\end{equation}
where $f(\mathbf{x}_{test})$ denotes the penultimate layer features of a test sample $\mathbf{x}_{test}$.
We then use the score $\mathcal{S}_{\text{TAP-Mahalanobis}}$ to distinguish between ID and OOD samples. To align with the conventional notion of having high score for ID samples and low score for OOD samples, negative sign is applied.

However, it is not straightforward to determine the value of the number of clusters $K$ for which the OOD detection performance of \emph{TAP-Mahalanobis} is optimal for different tasks and datasets. To illustrate the effect of using a different number of clusters $K$ in \emph{TAP-Mahalanobis} on the OOD detection performance, we conduct an ablation in \cref{sec:ablations}.
To this end, we present an \emph{Ensembling} module.\\

\begin{figure*}
\centering
\vspace{-2mm}
\begin{minipage}[t]{0.4\linewidth}
    \centering
    \includegraphics[width=0.75\linewidth]{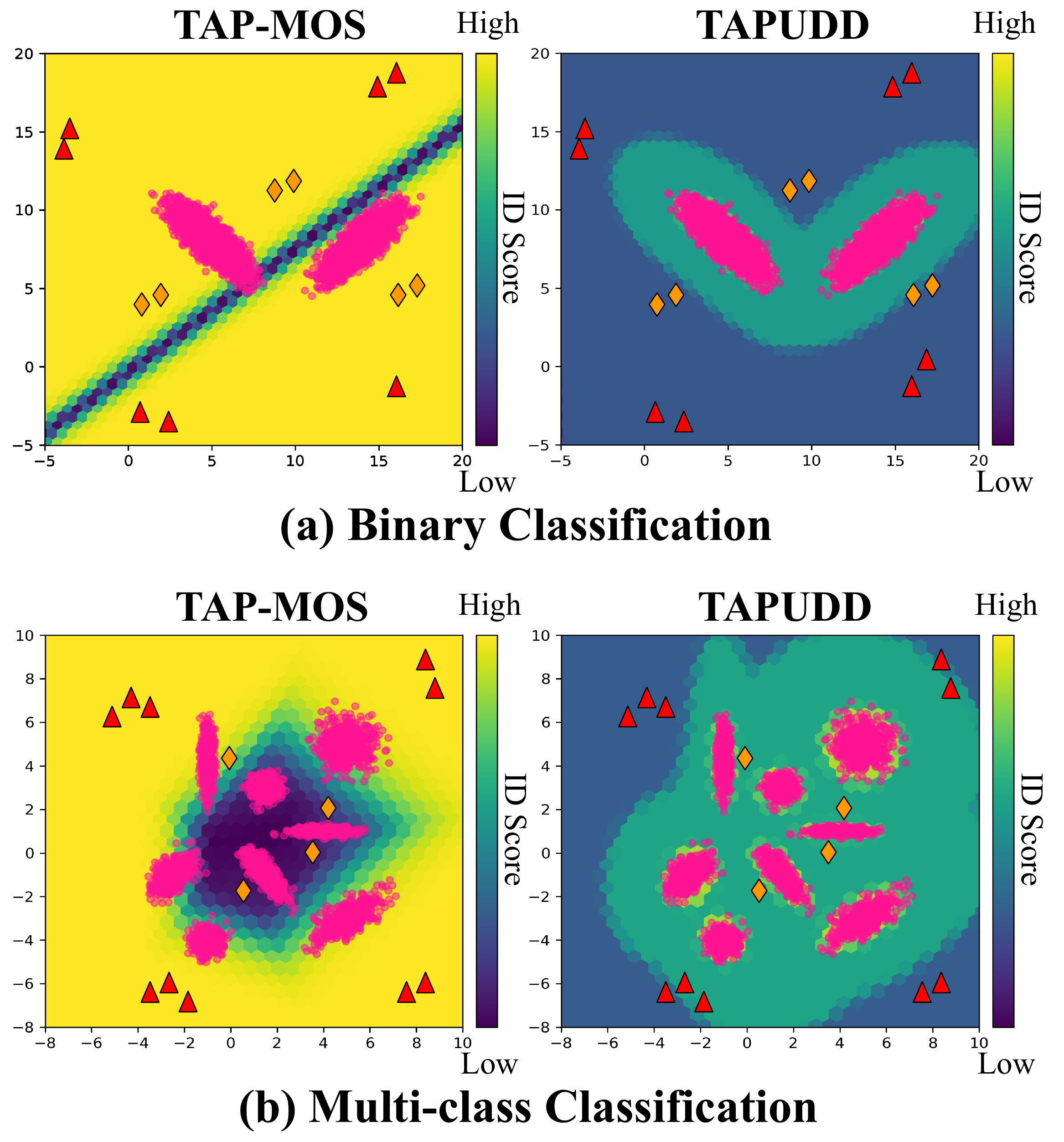}
    \caption{ID score landscape of TAP-MOS and TAPUDD on synthetic 2-D binary and multi-class classification datasets.
    A sample is \edit{deemed} as OOD when it has a
    \textcolor{blue}{\textbf{low ID score}}.
    The \textbf{\textcolor{magenta}{Pink Points}} represent the in-distribution data; \textbf{\textcolor{red}{Red Triangles}} and \textbf{\textcolor{orange}{Orange Diamonds}} represent the far and near OOD samples, respectively. 
    \edit{TAP-MOS fails to detect certain OOD samples, whereas TAPUDD effectively detects all OOD samples.}
    }
    \label{fig:synthetic_data_UMOS}
\end{minipage}%
\hfill
\begin{minipage}[t]{0.58\linewidth}
    \centering
    \includegraphics[width=0.98\linewidth]{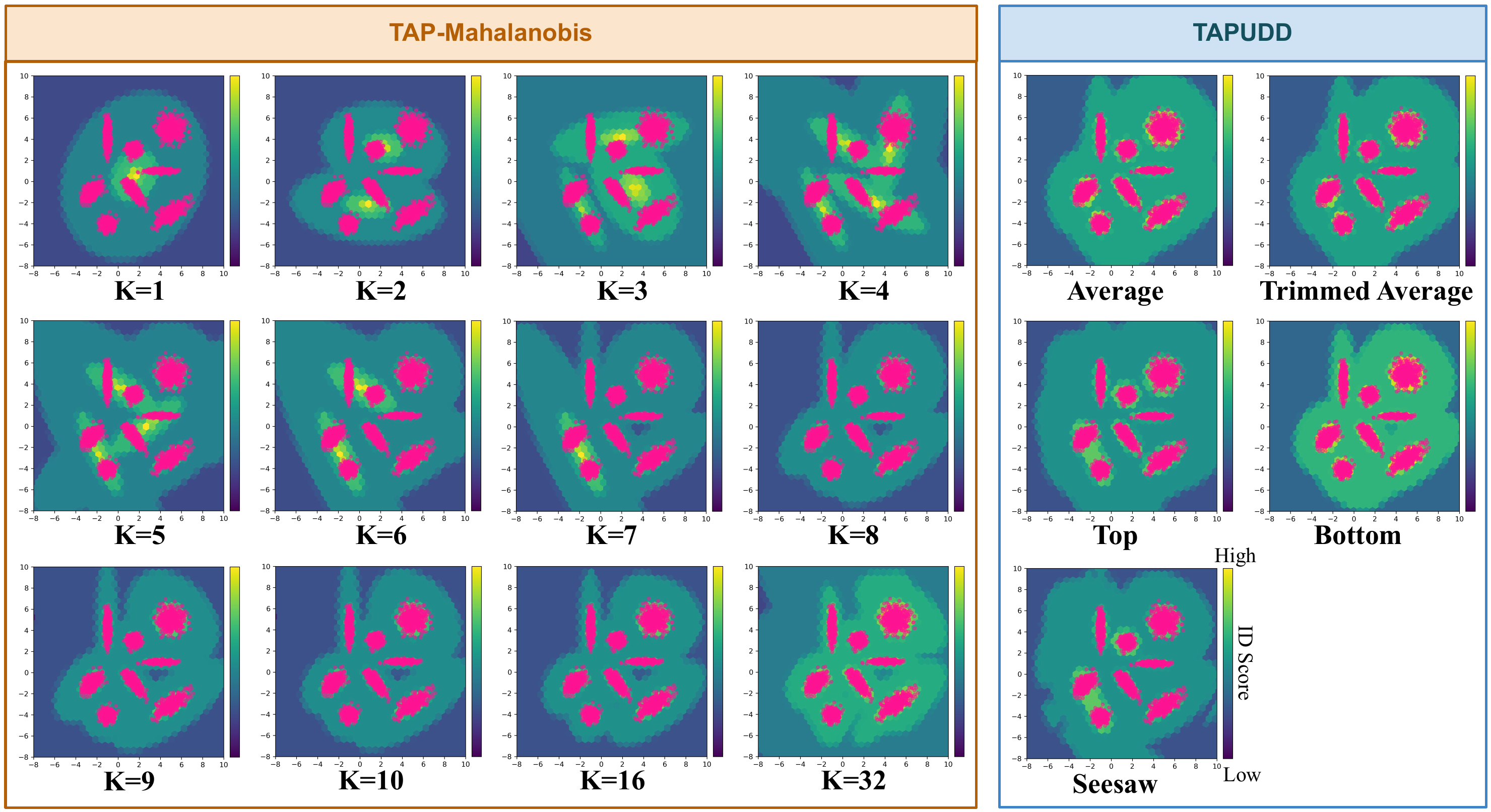}
    \caption{ID score landscape of \emph{TAP-Mahalanobis} for different values of $K$ (\ie , number of clusters); and TAPUDD for different ensemble variations on synthetic 2D multi-class classification dataset. A sample is deemed as OOD when it has a \textcolor{blue}{\textbf{low ID score}}. The \textbf{\textcolor{magenta}{Pink Points}} represent the in-distribution data.
    Results demonstrate that \emph{TAP-Mahalanobis} does not perform well for some values of $K$ whereas TAPUDD with all ensembling strategies perform better or on-par with \emph{TAP-Mahalanobis}. 
    }
    \label{fig:synthetic_data_umahala_vs_tauood}
\end{minipage}
\vspace{-1mm}
\end{figure*}

\vspace{-3mm}
\parahead{Ensembling} This module not only eliminates the need to determine the optimal value of $K$ but also provides more reliable results. \edit{We obtain \emph{TAP-Mahalanobis} scores for different values of $K \in [k_{1}, k_{2}, k_{3}, ..., k_{n}]$ and aggregate them to obtain an ensembled score, $\mathcal{S}_{\text{Ensemble}}$. This ensures that a sample is detected as OOD only if a majority of the participants in ensembling agrees with each other. }


\parahead{Remark} \editr{We empirically compare GMM with K-means and observe that GMM is more flexible in learning the cluster shape in contrast to K-means, which learned spherical cluster shapes. Consequently, K-means performs poorly when detecting OOD samples near the cluster.
Other popular clustering methods such as agglomerative clustering or DBSCAN are less compatible with Mahalanobis distance and require careful hyperparameter adjustment, such as the linking strategies for agglomerative clustering or the epsilon value for DBSCAN. 
Please refer to \cref{sec:discussion} for a detailed discussion on GMM clustering.
}

\subsection{Ensembling Strategies}
\edit{Given the dependency of TAPUDD on the aggregation of \emph{TAP-Mahalanobis} scores obtained for different values of $K$ (\ie , number of clusters)}
, a natural question arises: \textit{how do different ensembling strategies affect the performance of unseen distribution detection}? To answer this, we systematically consider the following five ensembling strategies:

\vspace{2pt}
\parahead{Average} We consider $\mathcal{S}_{\text{TAP-Mahalanobis}}$ obtained from \emph{TAP-Mahalanobis} module for different $K$ with equal importance and average them to obtain an 
ensembled score, $\mathcal{S}_{\text{Ensemble}}$.

\vspace{2pt}
\parahead{Trimmed Average} For certain values of $K$, the \emph{TAP-Mahalanobis} module can provide extremely high or extremely low $\mathcal{S}_{\text{TAP-Mahalanobis}}$. Therefore, to reduce bias caused by extreme \emph{TAP-Mahalanobis} scores, we eliminate \quotes{$m$} \emph{TAP-Mahalanobis} scores from top and bottom and then take the average of the remaining $n_e > K/2$ \emph{TAP-Mahalanobis} scores to obtain a final ensembled score.  

\vspace{2pt}
\parahead{Seesaw} In general, the voting ensemble in regression tasks includes the average of all participants. However, some participants might completely disagree with the other participants, and including all of them in ensembling might provide inaccurate results. 
To this end, we present \quotes{\textit{seesaw}} ensembling strategy wherein we sort the \emph{TAP-Mahalanobis} scores obtained for different values of $K$ and average the $n_e > K/2$ participants that agree with each other.
In other words, if a majority of participants agree to a high \emph{TAP-Mahalanobis} score, we pick the top $n_e$ participants; otherwise, we select the bottom $n_e$ participants. 

\vspace{2pt}
\parahead{Top} 
\edit{We sort the \emph{TAP-Mahalanobis} scores obtained for different values of $K$ and then obtain the ensembled score $\mathcal{S}_{\text{Ensemble}}$ by averaging the top $n_e > K/2$ scores.}

\vspace{2pt}
\parahead{Bottom}  
\edit{We sort the \emph{TAP-Mahalanobis} scores obtained for different values of $K$ and average the bottom $n_e > K/2$ scores to obtain an ensembled score $\mathcal{S}_{\text{Ensemble}}$.}

\section{Experiments and Results}
In this section, we validate our approach, TAPUDD, by conducting experiments on 2-D synthetic datasets 
(\cref{sec:synthetic_dataset_experiments}). To further bolster the effectiveness of our method, we present empirical evidence to validate TAPUDD on several real-world tasks, including binary classification  (\cref{sec:binary_classification_experiments}), regression (\cref{sec:regression_experiments}), and 
\edit{large-scale}
classification (\cref{sec:multi_classification_experiments}).  For binary classification and regression task, we evaluate our approach on Natural Attribute-based Shift (NAS) detection dataset. In NAS detection, a sample is shifted from the training distribution based on attributes like brightness, age, etc. Throughout all experiments, we use TAP-MOS, a task-agnostic and post-hoc extension of MOS\cite{Huang2021MOSTS}, as an additional baseline. Unlike the original MOS where class hierarchy was used to do clustering, we perform GMM clustering. 
\edit{More details on TAP-MOS are provided in \cref{sec:supp_tapmos}.}

\begin{table*}[t]
\vspace{-0.2mm}
\small
\centering
\resizebox{\linewidth}{!}{
\begin{tabular}{c  c  c  c c c  c c c c c }
\toprule
\multicolumn{1}{c}{\multirow{1}{*}{Brightness}} 
& \multicolumn{7}{c}{\multirow{1}{*}{Baselines}}  & \multicolumn{3}{c}{\multirow{1}{*}{Ours (Task-Agnostic)}} \\  \cmidrule(lr){2-8}  \cmidrule(lr){9-11} 
\multicolumn{1}{c}{} & \multicolumn{1}{c}{MSP~\cite{hendrycks2017base} } & \multicolumn{1}{c}{ODIN~\cite{liang2017enhancing} } & \multicolumn{1}{c}{Energy~\cite{liu2020energy} } & \multicolumn{1}{c}{MB~\cite{mahala18} } &  \multicolumn{1}{c}{KL~\cite{Hendrycks2019ABF} } &  \multicolumn{1}{c}{MOS~\cite{Huang2021MOSTS} } & \multicolumn{1}{c}{Gram~\cite{sastry2021gram}}  & \multicolumn{1}{c}{TAP-MOS} & \multicolumn{1}{c}{TAP-MB} & \multicolumn{1}{c}{TAPUDD} \\
& & & & & & (K = 8) & & (K = 8) & (K = 8) & (Average) 

\\ \midrule
        
0.0 & 88.7{\scriptsize $\pm$4.8} & 88.7{\scriptsize $\pm$4.8} & 88.2{\scriptsize $\pm$5.3} & 99.9{\scriptsize $\pm$0.1} & 26.3{\scriptsize $\pm$32.8} & 89.3{\scriptsize $\pm$5.5} & 99.3{\scriptsize $\pm$1.4} & 63.1{\scriptsize $\pm$26}   & 99.9{\scriptsize $\pm$0.1} & 100.0{\scriptsize $\pm$0.1} \\
0.2 & 66.1{\scriptsize $\pm$3.5} & 66.1{\scriptsize $\pm$3.5} & 66.0{\scriptsize $\pm$3.7}   & 87.5{\scriptsize $\pm$4.5} & 44.5{\scriptsize $\pm$3}    & 65.9{\scriptsize $\pm$3.2} & 61.0{\scriptsize $\pm$3.3}   & 63.6{\scriptsize $\pm$10}   & 86.8{\scriptsize $\pm$4.7} & 87.3{\scriptsize $\pm$5.2} \\
0.4 & 56.3{\scriptsize $\pm$1.4} & 56.4{\scriptsize $\pm$1.4} & 56.2{\scriptsize $\pm$1.7} & 70.5{\scriptsize $\pm$3.8} & 46.9{\scriptsize $\pm$1.2}  & 56.4{\scriptsize $\pm$1.1} & 53.4{\scriptsize $\pm$1.1} & 56.3{\scriptsize $\pm$4.9}  & 69.6{\scriptsize $\pm$3.7} & 70.1{\scriptsize $\pm$4.5}\\
0.6 & 52.4{\scriptsize $\pm$0.8} & 52.4{\scriptsize $\pm$0.8} & 52.3{\scriptsize $\pm$0.9} & 59.9{\scriptsize $\pm$2.5} & 48.2{\scriptsize $\pm$1}    & 52.5{\scriptsize $\pm$0.8} & 51.4{\scriptsize $\pm$0.5} & 52.7{\scriptsize $\pm$2.4}  & 59.3{\scriptsize $\pm$2.5} & 59.4{\scriptsize $\pm$2.7} \\
0.8 & 50.4{\scriptsize $\pm$0.4} & 50.4{\scriptsize $\pm$0.4} & 50.4{\scriptsize $\pm$0.4} & 52.2{\scriptsize $\pm$1.4} & 48.8{\scriptsize $\pm$0.6}  & 50.5{\scriptsize $\pm$0.3} & 50.2{\scriptsize $\pm$0.5} & 50.5{\scriptsize $\pm$1.2}  & 52.0{\scriptsize $\pm$1.7}   & 52.0{\scriptsize $\pm$1.6}   \\
 \rowcolor{Gray} 1.0 & 50.0{\scriptsize $\pm$0.0}     & 50.0{\scriptsize $\pm$0.0}     & 50.0{\scriptsize $\pm$0.0}     & 50.0{\scriptsize $\pm$0.0}     & 50.0{\scriptsize $\pm$0.0}      & 50.0{\scriptsize $\pm$0.0}     & 50.0{\scriptsize $\pm$0.0}     & 50.0{\scriptsize $\pm$0.0}      & 50.0{\scriptsize $\pm$0.0}     & 50.0{\scriptsize $\pm$0.0}     \\
1.2 & 51.7{\scriptsize $\pm$0.4} & 51.7{\scriptsize $\pm$0.4} & 51.7{\scriptsize $\pm$0.4} & 55.4{\scriptsize $\pm$1.6} & 49.2{\scriptsize $\pm$0.5}  & 51.7{\scriptsize $\pm$0.5} & 51.1{\scriptsize $\pm$0.6} & 51.2{\scriptsize $\pm$0.9}  & 56.1{\scriptsize $\pm$1.5} & 56.0{\scriptsize $\pm$1.5}    \\
1.4 & 55.8{\scriptsize $\pm$0.8} & 55.8{\scriptsize $\pm$0.8} & 55.8{\scriptsize $\pm$0.8} & 62.9{\scriptsize $\pm$2.1} & 48.2{\scriptsize $\pm$1.2}  & 55.8{\scriptsize $\pm$0.8} & 53.6{\scriptsize $\pm$1.1} & 53.6{\scriptsize $\pm$1.7}  & 63.7{\scriptsize $\pm$2.0}   & 63.5{\scriptsize $\pm$2.1} \\
1.6 & 59.7{\scriptsize $\pm$1.3} & 59.7{\scriptsize $\pm$1.3} & 59.8{\scriptsize $\pm$1.4} & 70.2{\scriptsize $\pm$2.7} & 47.5{\scriptsize $\pm$1.7}  & 59.6{\scriptsize $\pm$1.1} & 55.9{\scriptsize $\pm$1.2} & 55.8{\scriptsize $\pm$2.4}  & 70.9{\scriptsize $\pm$2.8} & 70.7{\scriptsize $\pm$2.9}   \\
1.8 & 63.1{\scriptsize $\pm$2.0}   & 63.1{\scriptsize $\pm$2.1} & 63.2{\scriptsize $\pm$2.2} & 76.5{\scriptsize $\pm$2.9} & 48.3{\scriptsize $\pm$2.4}  & 62.8{\scriptsize $\pm$1.7} & 58.1{\scriptsize $\pm$1.5} & 57.0{\scriptsize $\pm$4.0}      & 76.9{\scriptsize $\pm$3.4} & 76.6{\scriptsize $\pm$3.5} \\
2.0 & 65.5{\scriptsize $\pm$3.2} & 65.6{\scriptsize $\pm$3.2} & 65.7{\scriptsize $\pm$3.5} & 81.6{\scriptsize $\pm$2.7} & 49.8{\scriptsize $\pm$2.9}  & 65.1{\scriptsize $\pm$2.6} & 60.5{\scriptsize $\pm$1.8} & 57.7{\scriptsize $\pm$6.0}    & 81.8{\scriptsize $\pm$3.7} & 81.4{\scriptsize $\pm$3.8} \\
2.5 & 69.5{\scriptsize $\pm$6.5} & 69.5{\scriptsize $\pm$6.5} & 69.6{\scriptsize $\pm$6.8} & 90.4{\scriptsize $\pm$2.5} & 51.6{\scriptsize $\pm$4.9}  & 69.0{\scriptsize $\pm$5.5}   & 65.4{\scriptsize $\pm$4.4} & 60.0{\scriptsize $\pm$9.0}      & 89.9{\scriptsize $\pm$3.8} & 89.6{\scriptsize $\pm$4.1} \\
3.0 & 72.5{\scriptsize $\pm$8.7} & 72.5{\scriptsize $\pm$8.7} & 72.6{\scriptsize $\pm$9.0}   & 94.8{\scriptsize $\pm$1.8} & 51.3{\scriptsize $\pm$5.4}  & 72.0{\scriptsize $\pm$7.6}   & 69.6{\scriptsize $\pm$5.9} & 63.0{\scriptsize $\pm$11.5}   & 93.9{\scriptsize $\pm$3.8} & 93.6{\scriptsize $\pm$4.0}  \\
3.5 & 73.7{\scriptsize $\pm$9.7} & 73.7{\scriptsize $\pm$9.7} & 73.6{\scriptsize $\pm$10}  & 96.8{\scriptsize $\pm$1.3} & 52.0{\scriptsize $\pm$6.1}    & 73.0{\scriptsize $\pm$8.8}   & 72.2{\scriptsize $\pm$6.8} & 64.3{\scriptsize $\pm$13.6} & 95.5{\scriptsize $\pm$3.8} & 95.4{\scriptsize $\pm$3.7} \\
4.0 & 75.8{\scriptsize $\pm$9.5} & 75.8{\scriptsize $\pm$9.5} & 75.7{\scriptsize $\pm$9.8} & 97.8{\scriptsize $\pm$0.8} & 50.5{\scriptsize $\pm$7.2}  & 75.3{\scriptsize $\pm$8.8} & 75.1{\scriptsize $\pm$7.9} & 66.3{\scriptsize $\pm$14.4} & 96.5{\scriptsize $\pm$3.6} & 96.5{\scriptsize $\pm$3.2} \\
4.5 & 78.1{\scriptsize $\pm$7.9} & 78.1{\scriptsize $\pm$7.9} & 78.0{\scriptsize $\pm$8.3}   & 98.5{\scriptsize $\pm$0.5} & 47.4{\scriptsize $\pm$8.2}  & 77.8{\scriptsize $\pm$7.5} & 78.0{\scriptsize $\pm$7.1}   & 68.7{\scriptsize $\pm$13.7} & 97.3{\scriptsize $\pm$3.0}   & 97.4{\scriptsize $\pm$2.4}   \\
5.0 & 79.9{\scriptsize $\pm$6.4} & 79.9{\scriptsize $\pm$6.4} & 79.8{\scriptsize $\pm$6.9} & 98.8{\scriptsize $\pm$0.4} & 44.9{\scriptsize $\pm$8.4}  & 79.8{\scriptsize $\pm$6.1} & 80.4{\scriptsize $\pm$6.6} & 70.3{\scriptsize $\pm$12.8} & 97.9{\scriptsize $\pm$2.5} & 98.0{\scriptsize $\pm$1.7}    \\
5.5 & 81.4{\scriptsize $\pm$5.6} & 81.4{\scriptsize $\pm$5.6} & 81.3{\scriptsize $\pm$6.2} & 99.0{\scriptsize $\pm$0.4}   & 44.1{\scriptsize $\pm$8.7}  & 81.3{\scriptsize $\pm$5.4} & 82.4{\scriptsize $\pm$6.6} & 71.3{\scriptsize $\pm$12.7} & 98.2{\scriptsize $\pm$2.2} & 98.4{\scriptsize $\pm$1.2}   \\
6.0 & 82.5{\scriptsize $\pm$5.1} & 82.5{\scriptsize $\pm$5.1} & 82.5{\scriptsize $\pm$5.6} & 99.1{\scriptsize $\pm$0.4} & 43.6{\scriptsize $\pm$8.6}  & 82.4{\scriptsize $\pm$4.9} & 83.9{\scriptsize $\pm$6.3} & 71.9{\scriptsize $\pm$12.8} & 98.5{\scriptsize $\pm$1.9} & 98.7{\scriptsize $\pm$0.9}  \\
6.5 & 83.2{\scriptsize $\pm$4.9} & 83.2{\scriptsize $\pm$4.9} & 83.2{\scriptsize $\pm$5.4} & 99.2{\scriptsize $\pm$0.4} & 44.3{\scriptsize $\pm$8.2}  & 83.1{\scriptsize $\pm$4.6} & 85.0{\scriptsize $\pm$6.2}   & 72.2{\scriptsize $\pm$13.1} & 98.7{\scriptsize $\pm$1.7} & 98.9{\scriptsize $\pm$0.7}  \\
\midrule
Average & 67.8 & 67.8 & 67.8 & 82.1 & 46.9 & 67.7   & 66.8 & 61.0 & \textbf{81.7 }& \textbf{82.0 } \\
 \bottomrule
 \end{tabular}}
\vspace{0.1mm}
\caption{NAS detection performance in binary classification task for NAS shift of brightness in RSNA boneage dataset measured by AUROC. Highlighted row presents the performance on ID data. MB and TAP-MB refers to Mahalanobis and \emph{TAP-Mahalanobis}, respectively.
\edit{Our task-agnostic approach significantly outperforms all baselines (except MB) and is comparable to MB. Note that \textbf{MB is task-specific} and cannot be used in tasks other than classification.}
}
\label{tab:binary_class_nas}
\end{table*}

\subsection{Evaluation on Synthetic Datasets}
\label{sec:synthetic_dataset_experiments}

\parahead{Experimental Details} We generate synthetic datasets 
in $\mathbb{R}^{2}$ for binary classification and multi-class classification tasks. The in-distribution (ID) data $\mathbf{x}\in \mathcal{X} = \mathbb{R}^{2}$ is sampled from a Gaussian mixture model (refer to \cref{sec:supp_synthetic_data_gen} for more details). All the samples except the ID samples in the 2-D plane represent the OOD samples. We consider the 2-D sample as the penultimate layer features on which we can directly apply OOD detection methods like TAPUDD, \emph{TAP-Mahalanobis}, and TAP-MOS. 

\vspace{2pt}
\parahead{TAPUDD outperforms TAP-MOS}
We compare the OOD detection performance of TAP-MOS and TAPUDD. ~\cref{fig:synthetic_data_UMOS}\textcolor{red}{a} and ~\cref{fig:synthetic_data_UMOS}\textcolor{red}{b} presents the ID score landscape of TAP-MOS and TAPUDD for binary classification and multi-class classification, respectively. The \textbf{\textcolor{magenta}{Pink Points}} represent the in-distribution data; \textbf{\textcolor{red}{Red Triangles}} and \textbf{\textcolor{orange}{Orange Diamonds}} represent far and near OOD samples, respectively. Here for TAP-MOS, we present results using number of clusters as $2$ and $8$ in binary classification and multi-class classification, respectively. 
For TAPUDD, we present the results of the \quotes{\textit{average}} ensembling strategy computed across the number of clusters in $[1, 2, 3, 4, 5, 6, 7, 8, 9, 10, 16, 32]$.
We observe that our proposed approach TAPUDD detects all OOD samples effectively. We also notice that TAP-MOS fails to detect OOD samples
near or far from the periphery of all the clusters
.
Thus, to understand the scenarios where TAPMOS fails to identify OOD, we conduct a detailed analysis of the MOS approach in \cref{sec:supp_analysis} and observe that it also fails for the similar corner cases. 

\parahead{TAPUDD outperforms TAP-Mahalanobis}
We  present a comparison to demonstrate the effectiveness of TAPUDD against \emph{TAP-Mahalanobis} in \cref{fig:synthetic_data_umahala_vs_tauood}. 
We present the ID score landscape of \emph{TAP-Mahalanobis} for different values of $K$ and TAPUDD with different ensemble variations for multi-class classification in a 2-D synthetic dataset.
The \textbf{\textcolor{magenta}{Pink Points}} represent the in-distribution data. We observe that for certain values of $K$, \emph{TAP-Mahalanobis} fails to detect some OOD samples. However, all ensemble variations in TAPUDD effectively detect OOD samples and performs better, or on par, with \emph{TAP-Mahalanobis}. 
Thus, TAPUDD eliminates the necessity of choosing the optimal value of $K$.
We also provide results on a 2-D synthetic dataset for binary classification in \cref{sec:supp_additional_exp}.

\begin{table*}[t]
\vspace{-0.2mm}
\footnotesize
\centering
\resizebox{0.76\linewidth}{!}{
\begin{tabular}{c  c  c  c c c  c }
\toprule
\multicolumn{1}{c}{\multirow{1}{*}{Brightness}} 
& \multicolumn{3}{c}{\multirow{1}{*}{Baselines}}  & \multicolumn{3}{c}{\multirow{1}{*}{Ours (Task-Agnostic)}} \\  \cmidrule(lr){2-4}  \cmidrule(lr){5-7} 
\multicolumn{1}{c}{} & \multicolumn{1}{c}{DE~\cite{lakshminarayanan2017simple} } & \multicolumn{1}{c}{MC Dropout~\cite{gal2016dropout} } & \multicolumn{1}{c}{$\text{SWAG}^{*}$~\cite{maddox_2019_simple}}  & \multicolumn{1}{c}{TAP-MOS} & \multicolumn{1}{c}{TAP-MB} & \multicolumn{1}{c}{TAPUDD}\\
& & & & (K = 8) & (K = 8) & (Average) 

\\ \midrule
0.0 &  100.0{\scriptsize $\pm$NA} & 6.9{\scriptsize $\pm$NA}& 99.9{\scriptsize $\pm$NA} & 57.8{\scriptsize $\pm$31.5} & 100.0{\scriptsize $\pm$0.1} & 100.0{\scriptsize $\pm$0.0} \\
0.2 & 57.0{\scriptsize $\pm$NA} & 45.5{\scriptsize $\pm$NA} & 51.4{\scriptsize $\pm$NA} & 68.7{\scriptsize $\pm$18.0} & 87.9{\scriptsize $\pm$6.1} & 88.8{\scriptsize $\pm$6.7} \\
0.4 & 51.3{\scriptsize $\pm$NA} & 50.8{\scriptsize $\pm$NA} & 49.8{\scriptsize $\pm$NA} & 70.7{\scriptsize $\pm$16.4} & 64.5{\scriptsize $\pm$6.9} & 66.6{\scriptsize $\pm$5.0} \\
0.6 & 50.7{\scriptsize $\pm$NA} & 49.7{\scriptsize $\pm$NA} & 49.5{\scriptsize $\pm$NA} & 65.3{\scriptsize $\pm$11.5} & 54.6{\scriptsize $\pm$4.4} & 55.1{\scriptsize $\pm$2.5} \\
0.8 & 50.5{\scriptsize $\pm$NA} & 49.9{\scriptsize $\pm$NA} & 49.7{\scriptsize $\pm$NA} & 57.7{\scriptsize $\pm$6.0} & 48.9{\scriptsize $\pm$1.7} & 49.2{\scriptsize $\pm$1.0} \\
\rowcolor{Gray} 1.0 & 50.0{\scriptsize $\pm$NA} & 49.8{\scriptsize $\pm$NA} & 50.0{\scriptsize $\pm$NA} & 50.0{\scriptsize $\pm$0.0} & 50.0{\scriptsize $\pm$0.0} & 50.0{\scriptsize $\pm$0.0} \\
1.2 & 50.3{\scriptsize $\pm$NA} & 48.5{\scriptsize $\pm$NA} & 50.8{\scriptsize $\pm$NA} & 48.7{\scriptsize $\pm$4.0} & 57.6{\scriptsize $\pm$1.8} & 57.8{\scriptsize $\pm$1.9} \\
1.4 & 54.5{\scriptsize $\pm$NA} & 46.7{\scriptsize $\pm$NA} & 55.8{\scriptsize $\pm$NA} & 50.8{\scriptsize $\pm$8.0} & 68.4{\scriptsize $\pm$3.4} & 68.4{\scriptsize $\pm$3.4} \\
1.6 & 58.6{\scriptsize $\pm$NA} & 44.5{\scriptsize $\pm$NA} & 63.5{\scriptsize $\pm$NA} & 55.4{\scriptsize $\pm$11.3} & 78.7{\scriptsize $\pm$3.6} & 78.6{\scriptsize $\pm$3.7} \\
1.8 & 64.9{\scriptsize $\pm$NA} & 41.6{\scriptsize $\pm$NA} & 71.6{\scriptsize $\pm$NA} & 62.1{\scriptsize $\pm$14.5} & 86.4{\scriptsize $\pm$3.5} & 86.3{\scriptsize $\pm$3.6} \\
2.0 & 75.8{\scriptsize $\pm$NA} & 38.4{\scriptsize $\pm$NA} & 79.3{\scriptsize $\pm$NA} & 67.3{\scriptsize $\pm$16.8} & 91.9{\scriptsize $\pm$3.0} & 91.7{\scriptsize $\pm$3.2} \\
2.5 & 95.6{\scriptsize $\pm$NA} & 31.1{\scriptsize $\pm$NA} & 89.8{\scriptsize $\pm$NA} & 76.2{\scriptsize $\pm$16.4} & 97.5{\scriptsize $\pm$1.5} & 97.4{\scriptsize $\pm$1.4} \\
3.0 & 98.4{\scriptsize $\pm$NA} & 25.8{\scriptsize $\pm$NA} & 90.7{\scriptsize $\pm$NA} & 82.8{\scriptsize $\pm$13.5} & 99.0{\scriptsize $\pm$0.6} & 99.0{\scriptsize $\pm$0.5} \\
3.5 & 99.3{\scriptsize $\pm$NA} & 21.7{\scriptsize $\pm$NA} & 93.7{\scriptsize $\pm$NA} & 88.1{\scriptsize $\pm$10.2} & 99.4{\scriptsize $\pm$0.3} & 99.4{\scriptsize $\pm$0.3} \\
4.0 & 99.8{\scriptsize $\pm$NA} & 18.0{\scriptsize $\pm$NA} & 96.4{\scriptsize $\pm$NA} & 90.7{\scriptsize $\pm$6.8} & 99.6{\scriptsize $\pm$0.3} & 99.6{\scriptsize $\pm$0.2} \\
4.5 &  100.0{\scriptsize $\pm$NA} & 14.9{\scriptsize $\pm$NA} & 97.4{\scriptsize $\pm$NA} & 91.7{\scriptsize $\pm$4.4} & 99.7{\scriptsize $\pm$0.2} & 99.7{\scriptsize $\pm$0.1} \\
5.0 &  100.0{\scriptsize $\pm$NA} & 11.7{\scriptsize $\pm$NA} & 98.1{\scriptsize $\pm$NA} & 91.7{\scriptsize $\pm$3.8} & 99.8{\scriptsize $\pm$0.1} & 99.7{\scriptsize $\pm$0.1} \\
5.5 &  100.0{\scriptsize $\pm$NA} & 9.7{\scriptsize $\pm$NA}& 98.5{\scriptsize $\pm$NA} & 91.0{\scriptsize $\pm$4.5} & 99.8{\scriptsize $\pm$0.1} & 99.8{\scriptsize $\pm$0.2} \\
6.0 &  100.0{\scriptsize $\pm$NA} & 7.9{\scriptsize $\pm$NA}& 98.7{\scriptsize $\pm$NA} & 89.7{\scriptsize $\pm$5.7} & 99.8{\scriptsize $\pm$0.1} & 99.8{\scriptsize $\pm$0.2} \\
6.5 &  100.0{\scriptsize $\pm$NA} & 7.0{\scriptsize $\pm$NA}& 98.9{\scriptsize $\pm$NA} & 88.3{\scriptsize $\pm$7.0} & 99.8{\scriptsize $\pm$0.2} & 99.8{\scriptsize $\pm$0.3} \\
\midrule
Average & 77.8 & 31.0 & 76.7 & 72.2 & \textbf{84.2} & \textbf{84.3} \\
 \bottomrule
 \end{tabular}}
\vspace{1.4mm}
\caption{NAS detection performance in regression task (age prediction) for NAS shift of brightness in RSNA boneage dataset measured by AUROC. Highlighted row presents the performance on the ID dataset. DE, MC Dropout, TAP-MB, and NA denotes Deep Ensemble, Monte Carlo Dropout, \emph{TAP-Mahalanobis}, and Not Applicable respectively. $\text{SWAG}^{*}$ = SWAG + Deep Ensemble. 
}
\label{tab:regression_nas}

\end{table*}

\subsection{NAS Detection in Binary Classification}
\label{sec:binary_classification_experiments}

\parahead{Experimental Details}
\edit{We use the RSNA Bone Age dataset~\cite{013a4b4eeca44ecab48a66956acbb91d}, composed of left-hand X-ray images of the patient and their gender and age ($0$ to $20$ years). 
We alter the brightness of the X-ray images by a factor between $0$ and $6.5$ and form $20$ different NAS datasets to reflect the X-ray imaging set-ups in different hospitals following~\cite{Park2021NaturalAS}.
In-distribution data consists of images with a brightness factor $1.0$.
\edit{We trained a ResNet18 model using the cross-entropy loss and assessed it on the ID test set composed of images with a brightness factor of $1.0$.}
Further, we evaluate the NAS detection performance of our method and compare it with representative task-specific OOD detection methods on NAS datasets. Extensive details on the experimental set-up are described in \cref{sec:supp_experimental_details}.
For NAS detection, we measure the area under the receiver operating characteristic curve (AUROC), a commonly used metric for OOD detection. Additionally, we report the area under the precision-recall curve (AUPR) and the false positive rate of OOD examples when the true positive rate of in-distribution examples is at $95$\% (FPR95) in \cref{sec:supp_other_metrics}.}
 
\parahead{Results} The in-distribution classification accuracy
averaged across $10$ seeds 
of the gender classifier trained using cross-entropy loss
is $91.60$. We compare the NAS detection performance of our proposed approach with competitive post-hoc OOD detection methods in literature in \cref{tab:binary_class_nas}. \edit{As expected, the NAS detection performance of our approach and all baselines except KL Matching increase as the shift in the brightness attribute increases.}
We also observe that our proposed approaches, \emph{TAPUDD} and \emph{TAP-Mahalanobis} are more sensitive to NAS samples compared to competitive baselines, including Maximum Softmax Probability~\cite{hendrycks2017base}, ODIN~\cite{liang2017enhancing}, Mahalanobis distance~\cite{mahala18}, energy score~\cite{liu2020energy}, Gram matrices~\cite{sastry2021gram}, MOS~\cite{Huang2021MOSTS}, and  KL matching~\cite{Hendrycks2019ABF}. All these \edit{task-specific} baselines require the label information of the training dataset for OOD detection and cannot be used directly in tasks other than classification.
On the other hand, our proposed \edit{task-agnostic} approach does not require the access to class label information and it can be used across different tasks like regression. 

\subsection{NAS Detection in Regression}
\label{sec:regression_experiments}

\parahead{Experimental Details}
We use the RSNA Bone Age dataset (described in \cref{sec:binary_classification_experiments}) and solve the age prediction task. In this task, the objective is to automatically predict the patient's age given a hand X-ray image as an input.
\edit{As described in \cref{sec:binary_classification_experiments},}
we vary the brightness of images by a factor between $0$ and $6.5$ and form $20$ different NAS datasets.
In-distribution data comprises images of brightness factor $1.0$ (unmodified images).
We train a ResNet18 with MSE loss
and evaluate it on the test set composed of images with a brightness factor $1.0$.
Further, we evaluate the NAS detection performance of our proposed method and compare its performance with representative bayesian and non-bayesian uncertainty estimation methods on NAS datasets with attribute shift of brightness. Additionally, we compare the NAS detection performance of our approach with a well-known bayesian approach for uncertainty estimation, SWAG~\cite{maddox_2019_simple}. Extensive details on the experimental set-up are described in \cref{sec:supp_experimental_details}.
For NAS detection, we measure AUROC and additionally report AUPR and FPR95 in \cref{sec:supp_other_metrics}.

\begin{figure*}[hbt]
\vspace{-1.5mm}
\centering
     \includegraphics[width=0.82\textwidth]{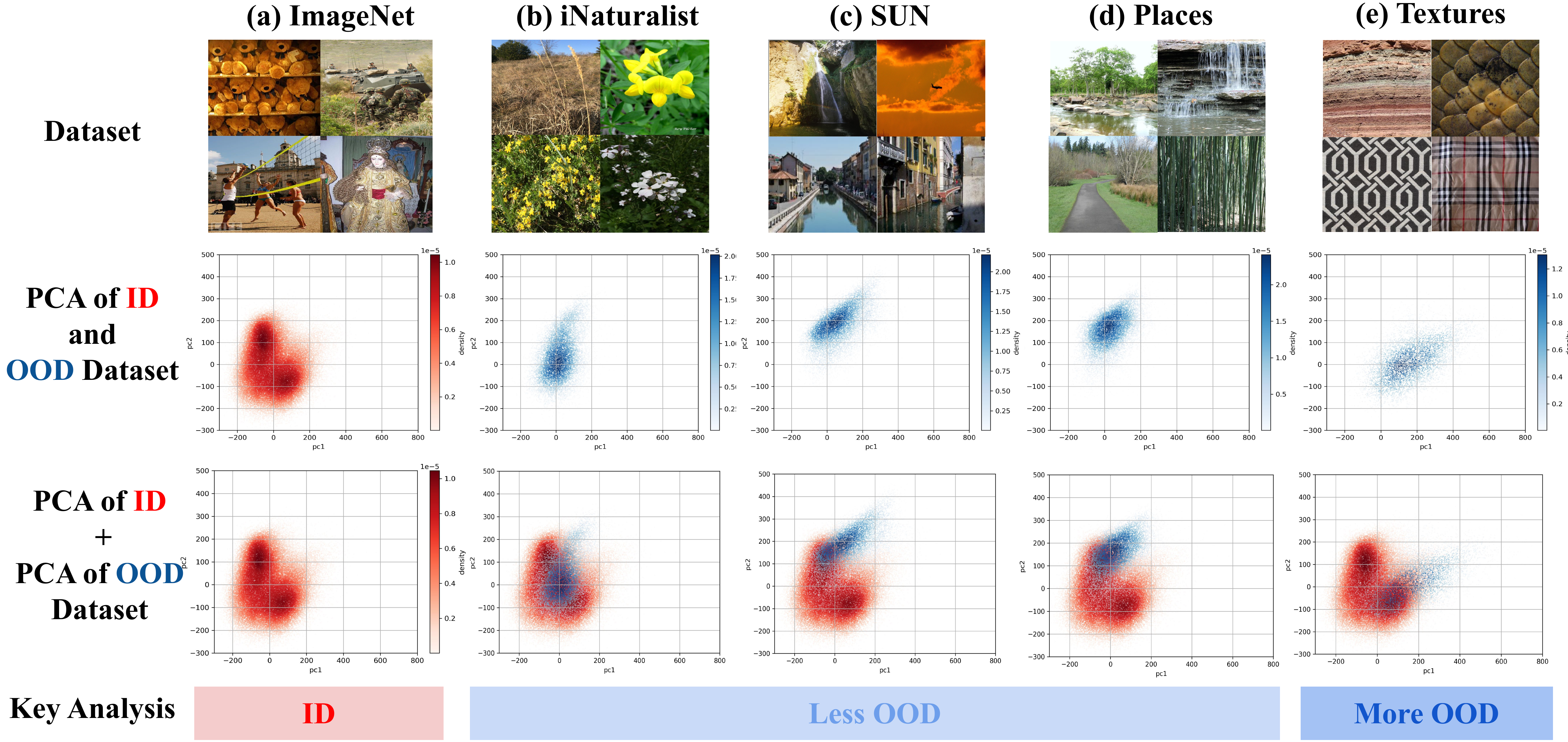}
     \caption{
     \edit{\emph{\textbf{(first row)}} Examples of ID images sampled from Imagenet and OOD images sampled from iNaturalist, SUN, Places, and Textures datasets; \emph{\textbf{(second row)}} Point-density based PCA visualization to demonstrate the location and density of ID and OOD datasets; \emph{\textbf{(third row)}} Point-density based PCA visualization of ID dataset overlapped by PCA of  OOD datasets to illustrate the location and density of OOD datasets relative to the ID dataset. \emph{\textbf{(fourth row)}} From \emph{\textbf{first}} and \emph{\textbf{third row}}, the key analysis is that \textbf{Textures is more OOD from Imagenet than the other three OOD datasets}.}}
     \label{fig:density_pca_ood_detection_final}
\end{figure*}

\begin{table*}[t]
\footnotesize
\centering
\resizebox{\linewidth}{!}{
\begin{tabular}{l|cc|cc|cc|cc|cc}
\toprule
\multirow{3}{*}{\textbf{Method}} & \multicolumn{2}{c|}{\textbf{iNaturalist}}    & \multicolumn{2}{c|}{\textbf{SUN}}            & \multicolumn{2}{c|}{\textbf{Places}}         & \multicolumn{2}{c|}{\textbf{Textures}}       & \multicolumn{2}{c}{\textbf{Average}}        \\ \cmidrule{2-11} 
                                 &                                                                                        \textbf{AUROC} $\uparrow$      & \textbf{FPR95} {$\downarrow$}       & \textbf{AUROC} $\uparrow$       & \textbf{FPR95} {$\downarrow$}         & \textbf{AUROC} $\uparrow$        & \textbf{FPR95} {$\downarrow$}         & \textbf{AUROC} $\uparrow$        & \textbf{FPR95} {$\downarrow$}         & \textbf{AUROC} $\uparrow$        & \textbf{FPR95} {$\downarrow$}        \\
                                 \midrule
\rowcolor{teagreen} \textbf{Expected Results} & Low & High & Low & High & Low & High & \textbf{High} & \textbf{Low}  & -- & --  \\ \midrule
MSP~\cite{hendrycks2017base} & 87.70 & 63.52 & 78.22 & 80.01 & 76.67 & 81.31 & 74.46 & 82.70 & 79.26 & 76.88 \\
ODIN~\cite{liang2017enhancing} & 89.49 & 62.61 & 83.83 & 71.89 & 80.60 & 76.51 & 76.29 & 81.31 & 82.55 & 73.08 \\
\rowcolor{teagreen}Mahalanobis \cite{mahala18} & 59.60 & 95.62 & 67.96 & 91.58 & 66.48 & 92.05 & 74.96 & 51.54 & 67.25 & 82.70 \\
Energy~\cite{liu2020energy} & 88.64 & 64.35 & 85.25 & 65.30 & 81.31 & 72.77 & 75.78 & 80.60 & 82.75 & 70.76 \\
KL Matching~\cite{Hendrycks2019ABF} & 93.06 & 27.24 & 78.74 & 67.56 & 76.53 & 72.67 & 87.07 & 49.47 & 83.85 & 54.23 \\
MOS~\cite{Huang2021MOSTS} & 98.15 & 9.23 & 92.01 & 40.38 & 89.05 & 49.49 & 81.27 & 60.30 & 90.12 & 39.85 \\ \midrule
\rowcolor{teagreen}\textbf{TAPUDD (Average)} & \textbf{70.00} & \textbf{84.46} & \textbf{70.47 }& \textbf{79.52} & \textbf{66.97} & \textbf{84.72} & \textbf{97.59} & \textbf{10.85} & \textbf{76.26} & \textbf{64.88}  \\
\bottomrule
\end{tabular}
}
\vspace{0.1mm}
\caption{
\edit{OOD detection performance in the large-scale classification task.
$\uparrow$ indicates larger values are better, while $\downarrow$ indicates smaller values are better. 
Ideally, all methods should follow the expected results obtained from our analysis (described in first row in \textcolor{ForestGreen}{green} color).
However, as highlighted in \textcolor{ForestGreen}{green} color, only Mahalanobis and our proposed approach follow the expected results. This highlights the failure of existing baselines, including MSP, ODIN, Energy, KL Matching, and MOS.
Further, amongst all methods following the expected results (highlighted in \textcolor{ForestGreen}{green} color), \textbf{\textit{our approach is highly sensitive to OOD samples and significantly outperforms the baselines}}. }
}

\label{table:multi_class_ood}
\end{table*}

\parahead{Results} The in-distribution Mean Absolute Error (MAE) in year averaged across $10$ seeds of the Resnet18 model trained using MSE loss is $0.801$. We compare the NAS detection performance of our proposed approach with well-known uncertainty estimation methods, namely  Deep Ensemble (DE) ~\cite{lakshminarayanan2017simple}, Monte Carlo Dropout (MC Dropout) ~\cite{gal2016dropout}, and SWAG~\cite{maddox_2019_simple}. Although Deep Ensemble, MC Dropout, and SWAG are not applicable to a pre-trained model, we compare against these baselines as a benchmark, as it has shown strong OOD detection performance across regression examples. For DE, we retrain $10$ models of the same architecture (Resnet18) using MSE loss from different initializations.  Since SWAG is not directly applicable for OOD detection, we apply  $\text{SWAG}^{*}$ which is a combination of deep ensembling on top of SWAG. 
From \cref{tab:regression_nas}, \redit{as expected, we observe that the NAS detection performance of our approach and all baselines increase as the shift in the brightness attribute increases.}
We also observe that our proposed approaches, TAPUDD and \emph{TAP-Mahalanobis}, are more sensitive to NAS samples and effectively detect them compared to the baselines and TAP-MOS. 
Additionally, it can be seen that TAP-MOS fails to detect extremely dark samples (Brightness Intensity $0.1$). This might be because these NAS samples could locate near or 
away from the periphery of all
clusters where MOS does not work well (as discussed in \cref{sec:synthetic_dataset_experiments}).
This demonstrates that our proposed approach is better than the existing approaches.

\begin{figure*}[t]
\centering
     \includegraphics[width=\textwidth]{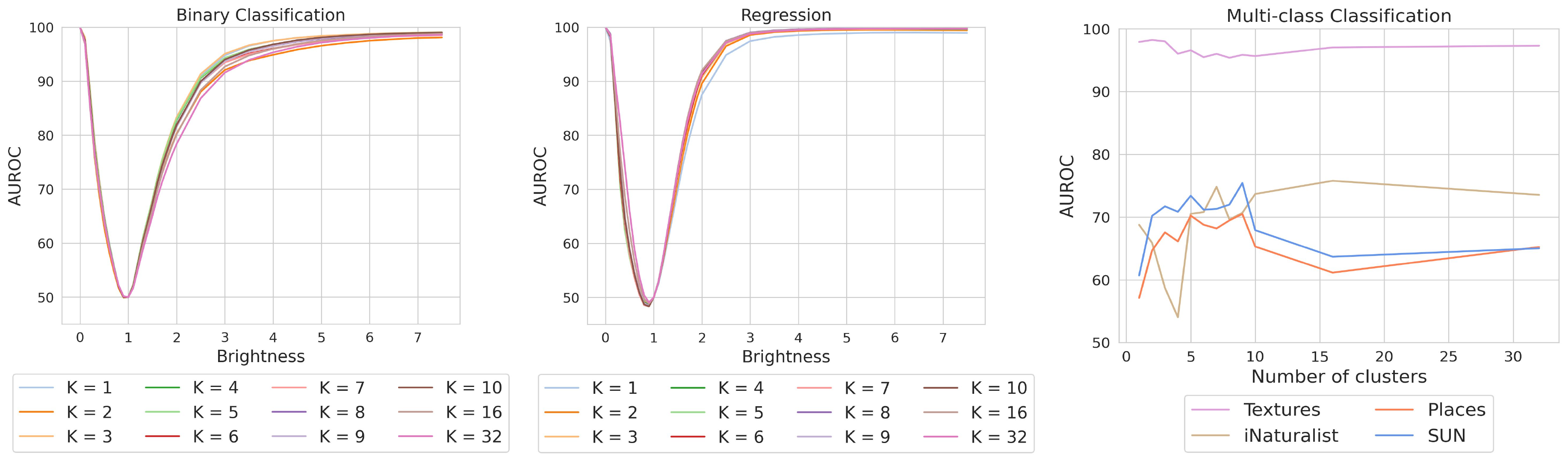}
     \vspace{-4mm}
     \caption{Effect on OOD detection performance by using different number of clusters in \emph{TAP-Mahalanobis}.}
     \label{fig:cluster_ablation}
     \vspace{2mm}
\end{figure*}

\begin{figure*}[t]
\centering
     \includegraphics[width=\textwidth]{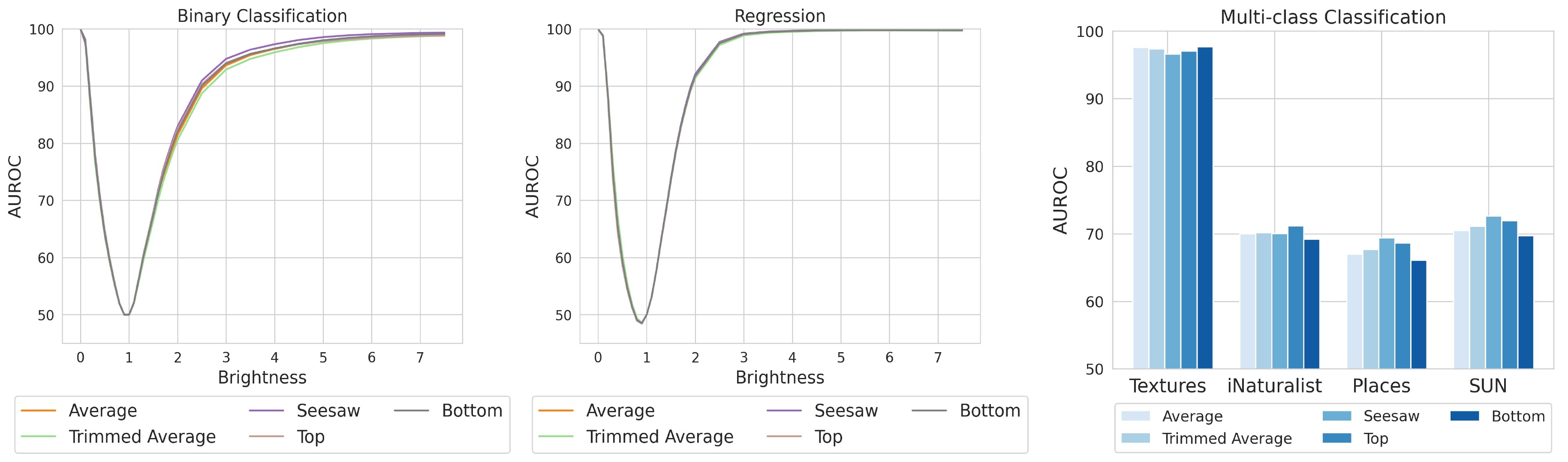}
     \vspace{-4mm}
     \caption{Comparison of OOD detection performance of TAPUDD under different ensembling strategies. 
     }
     \label{fig:ensemble_ablation}
\end{figure*}

\subsection{OOD Detection for Large-scale Classification}
\label{sec:multi_classification_experiments}

\parahead{Experimental Details}
We use ImageNet-1k\cite{Deng2009ImageNetAL}
as the in-distribution dataset and evaluate our approach on diverse OOD datasets presented in MOS\cite{Huang2021MOSTS}, including iNaturalist\cite{Horn2018TheIS}, Places\cite{Zhou2018PlacesA1}, SUN\cite{sun2014deep}, and Textures\cite{Cimpoi2014DescribingTI}. 
For all baselines, we follow the experimental setup used in MOS\cite{Huang2021MOSTS}.
We obtained the base model for our approach by following the experimental setting of finetuning the fully connected layer \redit{of pretrained ResNetv2-101 model} by flattened softmax used in MOS.
We measure AUROC and FPR95 and report AUPR in \cref{sec:supp_other_metrics}.

\parahead{Analysis}
\edit{In binary classification (\cref{sec:binary_classification_experiments}) and regression (\cref{sec:regression_experiments}) tasks, we evaluated our approach on NAS samples and knew that the NAS detection performance should increase with the increase in attribute shift.
However, in this experiment, since the shift does not increase in a continuous manner, 
it is non-trivial to determine
for which OOD datasets scores should be high.
To this end, we perform an analysis wherein
from the \emph{\textbf{first row}} of \cref{fig:density_pca_ood_detection_final}, we observe that the Textures dataset is visually much more different to human eyes from Imagenet compared to the other three OOD datasets. }
\edit{
Further, in the \emph{\textbf{second}} and \emph{\textbf{third rows}} of \cref{fig:density_pca_ood_detection_final}, we apply principal component analysis (PCA) on the feature representations obtained from the penultimate layer of the model to visualize the location of in-distribution samples (Imagenet) and diverse OOD datasets. From this, we observe that most of the samples from the Textures dataset are located away from the highly-dense regions of Imagenet compared to the other three datasets, thus indicating that Textures is more OOD. 
Hence, \textit{\textbf{we expect a reliable OOD detection model to detect Textures as more OOD than the other three OOD datasets}}. 
}

\parahead{Results}
\edit{To determine the OOD detection methods that follow the expected analysis, we evaluate the performance of our approach and well-known post-hoc OOD detection methods, including
Maximum Softmax Probability~\cite{hendrycks2017base}, ODIN~\cite{liang2017enhancing}, Mahalanobis distance ~\cite{mahala18}, energy score ~\cite{liu2020energy},  MOS~\cite{Huang2021MOSTS}, and  KL matching~\cite{Hendrycks2019ABF}. It is also worth noting that all the baselines require class label information for OOD detection and cannot be used for tasks other than classification. 
From \cref{table:multi_class_ood}, we observe that only our approach (TAPUDD) and Mahalanobis follow the expected results, as highlighted in \textcolor{ForestGreen}{green} color. This demonstrates that all baselines except Mahalanobis are less reliable.
Further, we compare the performance of methods following the expected results (rows highlighted in \textcolor{ForestGreen}{green} color) and observe that \textbf{\textit{our approach (TAPUDD) is more sensitive to OOD samples and outperforms the baselines in all OOD datasets}}.}

\subsection{Ablations}
\label{sec:ablations}

\parahead{Effect of Number of Clusters in TAP-Mahalanobis}
\edit{We investigate the effect of the number of clusters on the OOD detection performance of \emph{TAP-Mahalanobis} in \cref{fig:cluster_ablation}.}
We observe that the performance of \emph{TAP-Mahalanobis} varies with the value of K (\ie, number of clusters) across different datasets and tasks. This implies that we cannot use a particular value of K for all tasks and datasets.

\parahead{TAPUDD with Different Ensembling Strategies}
We contrast the OOD detection performance of TAPUDD under different ensembling strategies for the three tasks in \cref{fig:ensemble_ablation}. We observe that TAPUDD shows competitive performance with diverse ensembling strategies for all the tasks and dataset shifts. Also, we observe that \quotes{seesaw} is slightly more sensitive towards NAS samples in binary classification and regression, and towards OOD samples from Places and SUN in large-class classification. 
 
Further, \cref{fig:cluster_ablation} and \cref{fig:ensemble_ablation} illustrate that using ensembling strategies provides us with an OOD detector which is almost as good as the best performer of \emph{TAP-Mahalanobis}.

\edit{We also evaluate our approach on other conventional OOD datasets (ID: CIFAR-10; OOD: CIFAR-100, SVHN, etc) and for anomaly detection task in  \cref{sec:cifar_level} and \cref{sec:anomaly_detection}, respectively.}
\editr{Additionally, we provide a 
discussion on
tuning of hyperparameters in \cref{sec:discussion}.}

\section{Conclusion}
\label{sec:conclusion}
In this work, we propose a task-agnostic and post-hoc approach, TAPUDD, to detect samples from the unseen distribution. \editr{TAPUDD is a clustering-based ensembling approach 
composed of \emph{TAP-Mahalanobis} and \emph{Ensembling} modules.} \emph{TAP-Mahalanobis} module groups the semantically similar training samples into clusters
\editr{and determines the minimum Mahalanobis distance of the test sample from the clusters.}  
\edit{To enhance reliability and to eliminate the necessity to determine the optimal number of clusters for \emph{TAP-Mahalanobis}, the \emph{Ensembling} module aggregates the distances obtained from the \emph{TAP-Mahalanobis} module for different values of clusters.}
\editr{We validate the effectiveness of our approach by conducting extensive experiments on diverse datasets and tasks.}
\edit{As future work, it would be interesting to extensively evaluate TAPUDD to detect samples from unseen distribution in natural language processing, 3D vision, and healthcare.}

{\small
\bibliographystyle{wacv/template/ieee_fullname}
\bibliography{arxiv}
}

\clearpage
\newpage
\appendix

\begin{center}{\bf {\LARGE 
Appendix \\ [1em]
}
}
\end{center}

\parahead{Organization}
In the supplementary material, we provide:
\begin{itemize}
    \item a detailed description of TAP-MOS, a task-agnostic extension of MOS used as a baseline in most of our experiments (\cref{sec:supp_tapmos}).
    \item analysis of well-known OOD detection methods (\cref{sec:supp_analysis}).
    \item details on synthetic dataset generation (\cref{sec:supp_synthetic_data_gen}).
    \item \edit{discussion on the appropriateness of Mahalanobis distance, choice of GMM clustering in TAPUDD and tuning of hyperparameters (\cref{sec:discussion}).}
    \item extended description of datasets and experimental settings (\cref{sec:supp_experimental_details}).
    \item additional results on a synthetic dataset for binary classification, results on conventional OOD datasets for multi-class classification tasks, and results on anomaly detection task (\cref{sec:supp_additional_exp}).
    \item quantitative results with other metrics, including AUPR, FPR95 (\cref{sec:supp_other_metrics}).
\end{itemize}

\section{Task Agnostic and Post-hoc MOS (\emph{TAP-MOS})}
\label{sec:supp_tapmos}
We present an extension of MOS~\cite{Huang2021MOSTS} which was proposed for OOD detection in large-scale classification problems. 
Since we aim to present a baseline that does not rely on the label space, we develop a clustering-based OOD detection method, Task Agnostic and Post-hoc Minimum Others Score (\emph{TAP-MOS}), in the features space. 
The training datasets' features extracted from a model trained for a specific task are given as input to the \emph{TAP-MOS}  module.
\emph{TAP-MOS}  module partition the features of in-distribution data into $K$ clusters using Gaussian Mixture Model (GMM) with \quotes{\textit{full}} covariance and train a cluster classification model. Motivated by the success of MOS, we perform group based learning and form $K$ groups, $\mathcal{G}_\text{1}$, $\mathcal{G}_\text{2}$, ..., $\mathcal{G}_\text{K}$,  where each group $\mathcal{G}_\text{k}$ comprises of samples of cluster $k$. A new category \quotes{\textit{others}} is then introduced in each group $\mathcal{G}_\text{k}$. 
The class labels in each group are re-assigned during the training of cluster classification model. \quotes{\textit{Others}} class is defined as the ground-truth class for the groups that do not include cluster $c$.
Following MOS~\cite{Huang2021MOSTS}, we calculate the group-wise softmax for each group $\mathcal{G}_\text{k}$ as:
\begin{equation}
    p_{c}^k(\mathbf{x}) = \frac{e^{f_{c}^{k}(\mathbf{x})}}{\sum_{c^{'}\in \mathcal{G}_{\text{i}} } e^{f_{c^{'}}^{k}(\mathbf{x})}}, c \in \mathcal{G}_{\text{k}},
    \label{eq:group_softmax}
\end{equation}
where $f_{c}^{k}(\mathbf{x})$ and $p_{c}^{k}(\mathbf{x})$ represent the output logit and softmax probability of a class $c$ in group $\mathcal{G}_\text{k}$, respectively.

The training objective for cluster classification is the sum of cross-entropy losses across all the groups:
\begin{equation}
    \mathcal{L} = -\frac{1}{N} \sum_{n=1}^{N} \sum_{k=1}^{K} \sum_{c \in \mathcal{G}_{\text{k}}} y_{c}^{k} \log{(p_{c}^{k}(\mathbf{x}))},
    \label{eq:mos_loss}
\end{equation}
where $y_{c}^{k}$ and $p_{c}^{k}(\mathbf{x})$ denote the re-assigned class labels and the softmax probability of category $c$ in $\mathcal{G}_{k}$, and $N$ denotes the total number of training samples.

For OOD detection, we utilize the \textit{Minimum Others Score} (MOS) which uses the lowest \quotes{\textit{others}} score among all groups to distinguish between ID and OOD samples and is defined as:
\begin{equation}
    \mathcal{S}_{\text{\emph{TAP-MOS} }} = - \min\limits_{1 \leq k \leq K} p_{others}^{k}(\mathbf{x}).
    \label{eq:mos_Score}
\end{equation}
To align with the conventional notion of having high score for ID samples and low score for OOD samples, the negative sign is applied.
We hypothesize that \emph{TAP-MOS} will fail when samples are near or away from the periphery of all clusters because it performs One-vs-All classification and detect a sample as OOD only if it is detected as \quotes{\textit{others}} class by all groups. We validate our hypothesis by conducting experiments on the synthetic datasets in Section 4.1 of the main paper. 
\section{Analysis of Well-known OOD Detection Methods}
\label{sec:supp_analysis}

\begin{figure*}[t]
\centering
     \includegraphics[width=1.0\textwidth]{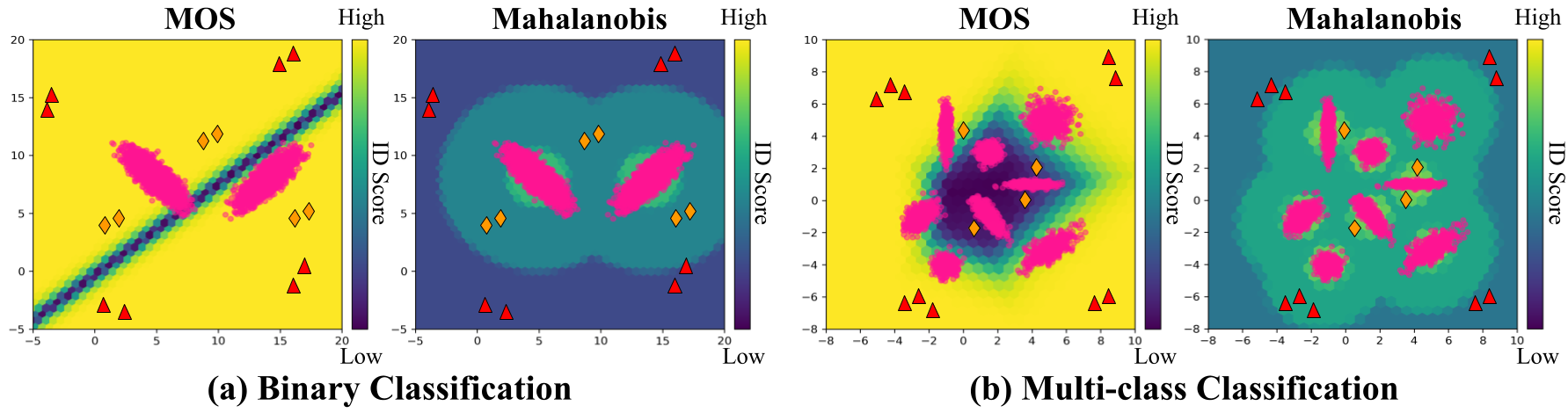}
     \vspace{-4mm}
     \caption{\small{ID score landscape of the existing representative post-hoc OOD detection methods (Mahalanobis, and MOS) on synthetic 2D binary and multi-class classification datasets. A sample is regarded as OOD when it has a \textcolor{blue}{\textbf{low ID score}}. The \textbf{\textcolor{magenta}{Pink Points}} represent the in-distribution data; \textbf{\textcolor{red}{Red Triangles}} and \textbf{\textcolor{orange}{Orange Diamonds}} represents OOD samples. 
    Results demonstrate that MOS fails to detect
    samples near or away from the periphery of all classes (\eg , \textbf{\textcolor{red}{Red Triangles}}) and Mahalanobis fails to detect samples near the ID classes (\eg ,  \textbf{\textcolor{orange}{Orange Diamonds}}) as OOD. }}
    \label{fig:mos_mahala_failure}
\end{figure*}

\parahead{MOS} MOS~\cite{Huang2021MOSTS} performs One-vs-All classification and detects a sample as OOD only if it is detected as the \quotes{\textit{others}} class by all groups. We hypothesize that MOS will fail to detect samples near or away from the periphery of all classes as OOD. This is because the groups in the corner will detect samples on the same side of the decision boundary of the one vs. all groups classification as ID. We now conduct an experiment on the 2D synthetic datasets for binary and multi-class classification tasks to determine if our hypothesis holds true. More details on synthetic datasets are provided in
\cref{sec:supp_synthetic_data_gen}. ~\cref{fig:mos_mahala_failure}\textcolor{red}{a} and ~\cref{fig:mos_mahala_failure}\textcolor{red}{b} presents the ID score landscape of MOS in binary class and multi-class classification tasks respectively. The \textbf{\textcolor{magenta}{Pink Points}} represent the in-distribution data; \textbf{\textcolor{red}{Red Triangles}} and \textbf{\textcolor{orange}{Orange Diamonds}} represents OOD samples. Results demonstrate that MOS works very well when the OOD samples are in between multiple classes (can be seen in blue color in 2-D plane). However, it fails to detect samples near or away from the corner classes (denoted by \textbf{\textcolor{red}{Red Triangles}}).

\parahead{Mahalanobis}
Mahalanobis OOD detector~\cite{mahala18} approximates each class as multi-variate gaussian distribution with tied covariance. However, in reality, the features can be correlated differently in different classes. In particular, features of a few classes can be correlated positively, and features of some other classes might be correlated negatively. We hypothesize that Mahalanobis might fail to detect OOD samples near the ID classes in such cases.
We conduct an experiment
in the same manner as above to test our hypothesis.
~\cref{fig:mos_mahala_failure}\textcolor{red}{a} and ~\cref{fig:mos_mahala_failure}\textcolor{red}{b} presents the ID score landscape of Mahalanobis in binary class and multi-class classification tasks respectively.
Results demonstrate that Mahalanobis works very well when the OOD samples are located far away from the ID classes but it fails to detect samples located near the ID classes (denoted by \textbf{\textcolor{orange}{Orange Diamonds}}).

\parahead{SSD}
SSD~\cite{sehwag2021ssd} is an outlier detector based on unlabeled in-distribution data which utilizes self-supervised representation learning followed by Mahalanobis distance based OOD detection. In self-supervised representation learning, SSD uses $\tiny{NT-Xent}$ loss function from SimCLR ~\cite{Chen2020ASF} which utilizes multiple data augmentation techniques.
SSD has achieved remarkable OOD detection performance and even outperforms several supervised OOD detection methods. 
\begin{figure}[hbt]
\centering
     \includegraphics[width=\columnwidth]{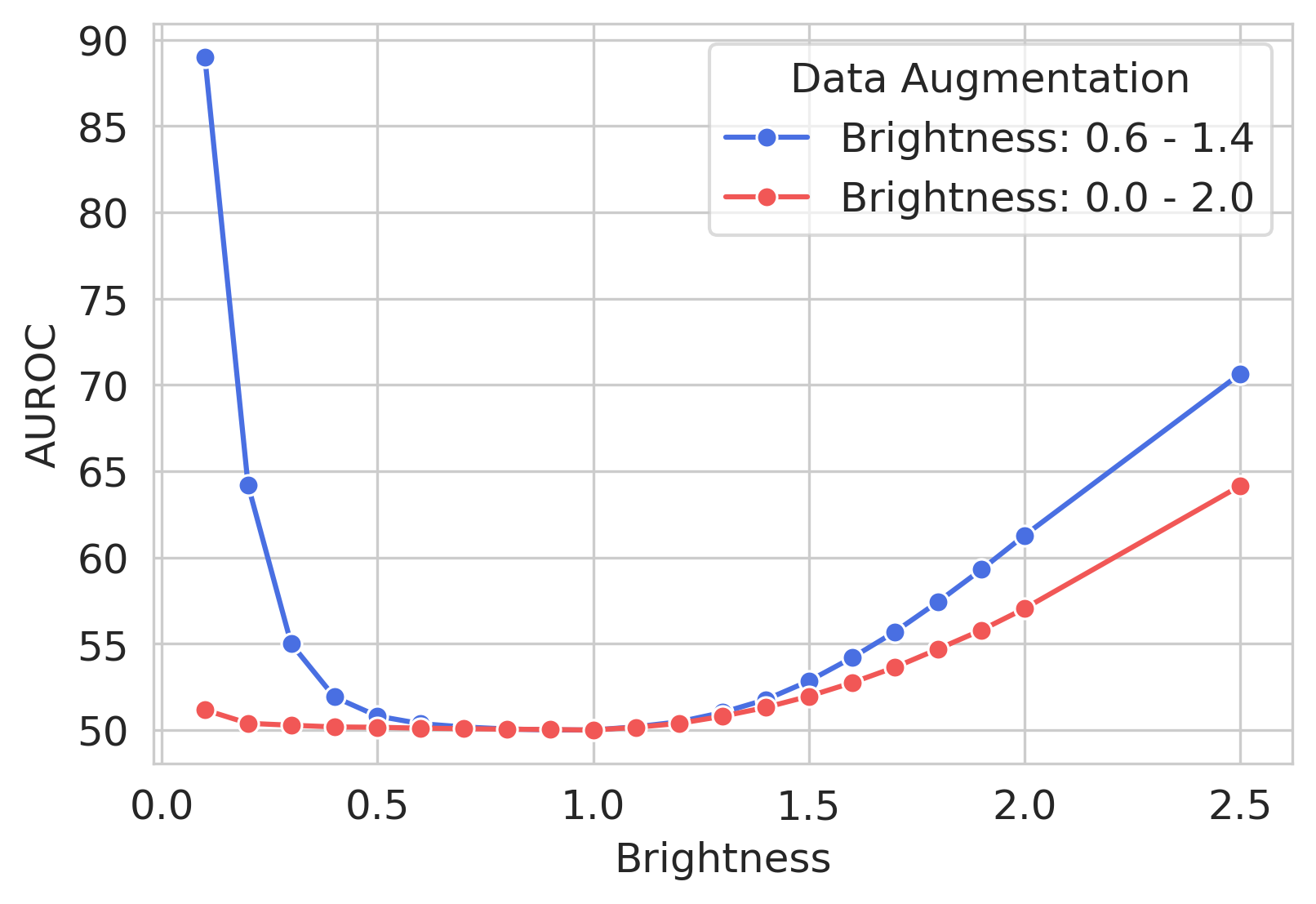}
     \caption{\small{Impact of varying intensity of brightness in data augmentation on the OOD detection performance of SSD.}
     }
     \label{fig:ssd-failure}
\end{figure}
However, we hypothesize that when OOD samples are from the same distribution as the augmented data, the model might fail to detect them as OOD. 
This can be problematic in several real-world tasks. For instance, in 3D vision, it is desirable to detect the shapes rotated by more than some threshold. However, if we used the rotation technique in data augmentation, SSD might fail to detect the samples as OOD. This can also lead to severe consequences in safety-critical applications. 
We conduct an experiment to determine if our hypothesis holds true. We used the CIFAR-10 dataset for self-supervised training in SSD and used data augmentation similar to SSD and varied the range of intensity of 
brightness used in data augmentation. 
Then, we evaluate the SSD model in the NAS setup. More specifically, we shift the
brightness of CIFAR-10 samples with varying levels of intensities and evaluate the performance of SSD 
when trained with augmentation of brightness.
~\cref{fig:ssd-failure} presents the impact of using different intensities of brightness in data augmentation on the performance of SSD for NAS detection. We observe that when the brightness intensity from 0.6 to 1.4 is used 
in data augmentation, the SSD model fails to detect samples from these brightness intensities as OOD. Further, when the brightness intensity from 0.0 to 2.0 is used in data augmentation, the SSD model even fails to detect extraordinarily dark and light images. This demonstrates that SSD fails to detect OOD samples from the same distribution as the augmented data. Similar to SSD, OOD detection methods that utilizes data augmentation for self-supervised learning might fail in scenarios where the model encounters OOD samples from distribution same as the distribution of augmented dataset. Therefore, we do not compare our approach against such OOD detection methods in all the experiments.

We observe that MOS, Mahalanobis, and SSD do not perform well in certain scenarios.
Moreover, MOS and Mahalanobis OOD detection methods require the class label information of the training datasets. Therefore, they cannot be directly used for OOD detection in tasks other than classification. This motivates the necessity of an unsupervised  OOD detection method that is not only task-agnostic and architecture-agnostic but also addresses the scenarios where MOS, Mahalanobis, and SSD do not perform well.

\section{Synthetic Dataset Generation}
\label{sec:supp_synthetic_data_gen}
\begin{figure}[hbt]
\centering
    \vspace{-2mm}
     \includegraphics[width=\columnwidth]{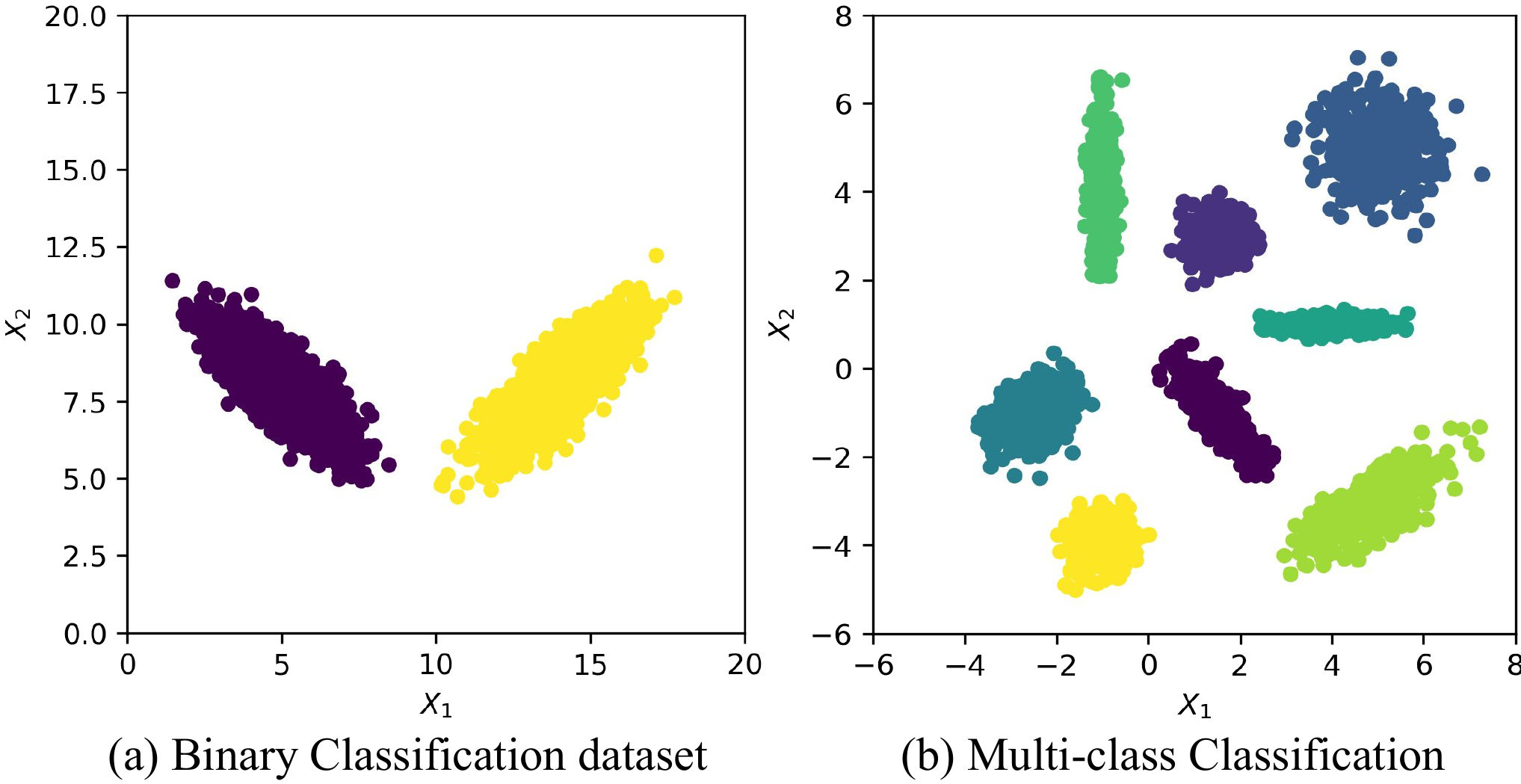}
     \caption{\small{2-D synthetic datasets for (a) binary classification and (b) multi-class classification tasks. }}
    \label{fig:synthetic_data}
\end{figure}
We generate synthetic datasets in $\mathbb{R}^{2}$ for binary and multi-class classification tasks, as shown in \cref{fig:synthetic_data}. The in-distribution (ID) data $\mathbf{x}\in \mathcal{X} = \mathbb{R}^{2}$ is sampled from a Gaussian mixture model. All the samples except the ID samples in the 2-D plane represent the OOD samples.
Dataset for binary classification and multi-class classification tasks comprises of $2$ and $8$ clusters, respectively. For each cluster with mean ($\mathcal{\mu}_{x}$, $\mathcal{\mu}_{y}$) and covariance, $3000$ and $500$ data points are sampled in binary and multi-class classification tasks, respectively. More details on the mean and covariance of each cluster is provided in \cref{tab:supp_synthetic_dataset}.

\begin{table}[h]
\vspace{2mm}
\scriptsize
\centering
\resizebox{\columnwidth}{!}{
\begin{tabular}{c c | c | c }
\toprule
Dataset & Cluster & Mean ($\mathcal{\mu}_{x}$, $\mathcal{\mu}_{y}$) & Covariance \\ \midrule
\multirow{2}{*}{Binary } & Cluster $1$ & ($5.0$, $8.0$) & ([$1.0$, -$0.8$], [-$0.8$, $1.0$])\\
& Cluster $2$  & ($14.0$, $8.0$) & ([$1.0$, $0.8$], [$0.8$, $1.0$]) \\
\midrule
\multirow{8}{*}{Multi-class}  & Cluster $1$ & ($1.5$, -$1.0$) & ([$0.2$, -$0.3$], [-$0.2$, $0.2$]) \\
& Cluster $2$ & ($1.5$, $3.0$) & ([$0.1$, $0.0$], [$0.0$, $0.1$]) \\ 
& Cluster $3$ & ($5.0$, $5.0$) & ([$0.4$, $0.0$], [$0.0$, $0.4$]) \\ 
& Cluster $4$ & (-$2.5$, -$1.0$) & ([$0.1$, $0.2$], [$0.2$, $0.1$]) \\ 
& Cluster $5$ & ($4.0$, $1.0$) & ([$0.4$, $0.0$], [$0.0$, $0.01$]) \\ 
& Cluster $6$ & (-$1.0$, $4.3$) & ([$0.02$, $0.0$], [$0.0$, $0.7$]) \\ 
& Cluster $7$ & ($5.0$, -$3.0$) & ([$0.5$, $0.4$], [$0.3$, $0.1$]) \\  
& Cluster $8$ & (-$1.0$, -$4.0$) & ([$0.1$, $0.0$], [$0.0$, $0.1$]) \\ 
 \bottomrule
 \end{tabular}}
 \vspace{0.5mm}
\caption{Mean and covariance per cluster used for generating dataset for binary and multi-class classification tasks.
}
\label{tab:supp_synthetic_dataset}
\end{table}

\begin{figure*}[t]
\centering
     \includegraphics[width=0.92\textwidth]{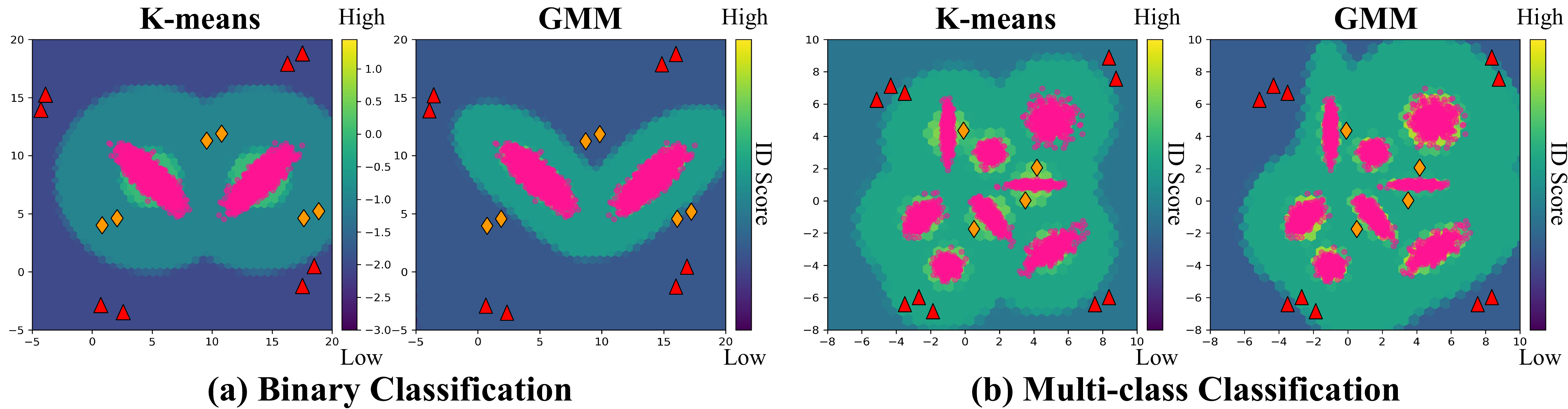}
     \caption{\small{ID score landscape of TAP-Mahalanobis on synthetic 2D binary and multi-class classification datasets on using K-means and GMM with full covariance. A sample is deemed as OOD when it has a \textcolor{blue}{\textbf{low ID score}}. The \textbf{\textcolor{magenta}{Pink Points}} represent the in-distribution data; \textbf{\textcolor{red}{Red Triangles}} and \textbf{\textcolor{orange}{Orange Diamonds}} represent OOD samples. 
    Results demonstrate that on using K-means, TAP-Mahalanobis fails to detect OOD
    samples near the clusters (\eg ,  \textbf{\textcolor{orange}{Orange Diamonds}}). However, on using GMM with full covariance, Mahalanobis effectively detects all OOD samples. }}
    \label{fig:synthetic_data_kmeans}
\end{figure*}

\begin{figure*}[hbt]
\centering
     \includegraphics[width=0.9\textwidth]{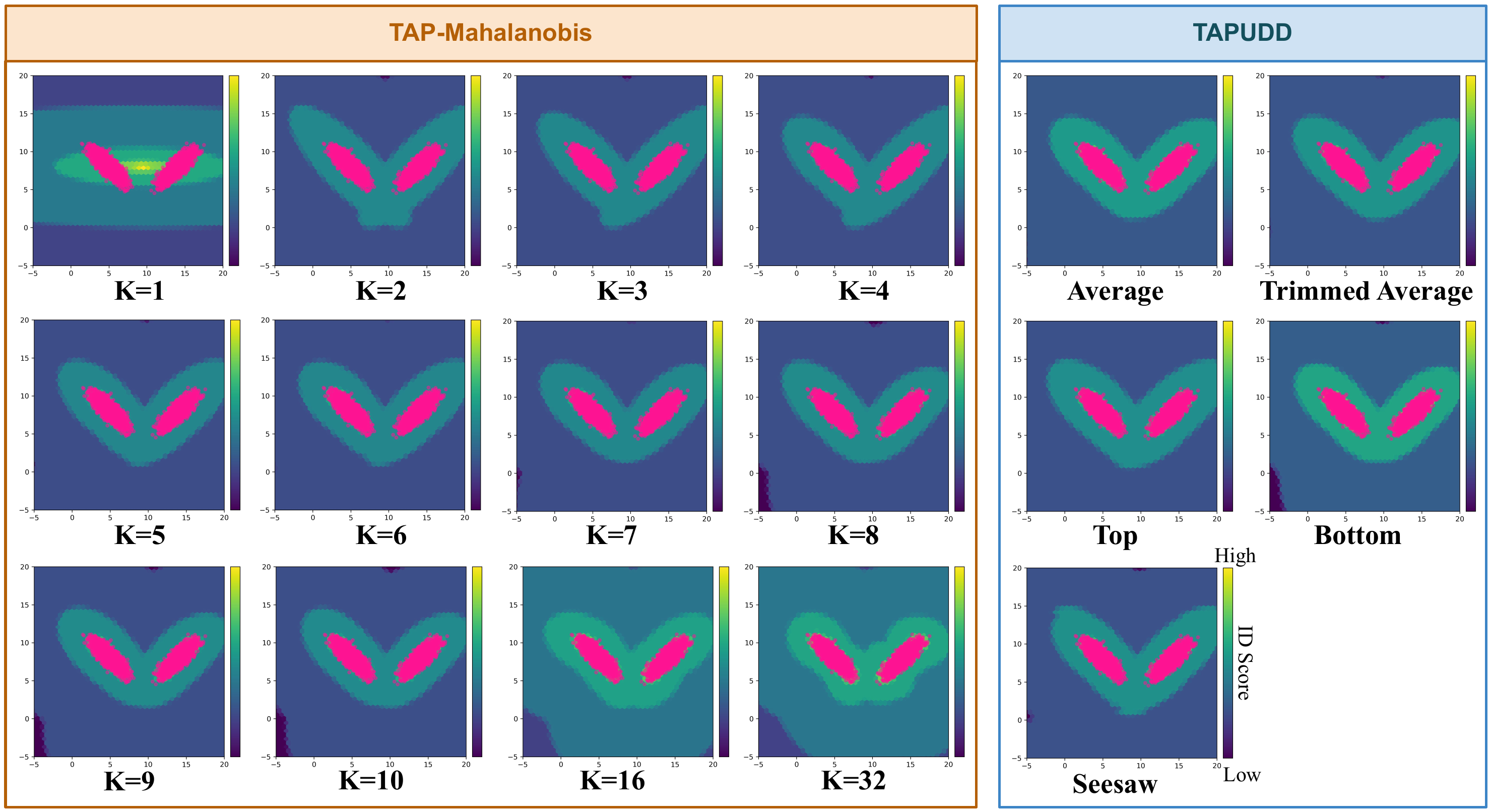}
     \caption{\small{ID score landscape of TAP-Mahalanobis for different values of $K$ (\ie , number of clusters); and TAPUDD for different ensemble variations on synthetic 2D binary classification dataset. A sample is regarded as OOD when it has a \textcolor{blue}{\textbf{low ID score}}. The \textbf{\textcolor{magenta}{Pink Points}} represent the in-distribution data.
    Results demonstrate that TAP-Mahalanobis does not perform well for some values of $K$ whereas TAPUDD with all ensembling strategies perform better or on-par with TAP-Mahalanobis. 
    }}
    \label{fig:synthetic_data_umahala_vs_tauood_binary}
\end{figure*}

\section{Discussion}
\label{sec:discussion}

\parahead{Appropriateness of Mahalanobis distance (MD)} \editr{Given that density estimation in high-dimensional space is a known intractable problem, we view MD as a reasonable approximation that leads to empirical efficacy.
Moreover, we believe that MD in TAPUDD is safe since the ensembling module does a \textbf{\textit{reasonably simple approximation}} by aggregating the MD obtained from GMMs with a different number of clusters. 
Evaluating the compatibility of our test-time framework on methods trained with added regularization to explicitly make Mahalanobis distance more appropriate for OOD detection~\cite{liu2017incremental} can be an interesting future direction to explore. }

\parahead{Reason for using GMM} \edit{We aim to use Mahalanobis distance to measure the distance between a test sample and local training set clusters in the latent space, hence GMM with full covariance is a natural fit. 
We compare GMM with K-means in \cref{fig:synthetic_data_kmeans} and observe that GMM is flexible in learning the cluster shape in contrast to K-means, which learned spherical cluster shapes. Consequently, K-means performs poorly when detecting OOD samples near the cluster.
Other popular clustering methods such as agglomerative clustering or DBSCAN are less compatible with Mahalanobis distance and require careful hyperparameter adjustment, such as the linking strategies for agglomerative clustering or the epsilon value for DBSCAN.}

\parahead{Tuning of Hyperparameters} \edit{Although it is possible to tune hyperparameters $K$ and $n_e$, our experiments indicate there is very little need to tune them. We observe that our approach can effectively detect OOD samples across different tasks and datasets as long as  $K$ consists of a sufficient number of diverse clusters (approximately $12$) and $n_e$ is equal to more than half of the number of participants in $K$.
We used $K = [1, 2, 3, 4, 5, 6, 7, 8, 9, 10, 16, 32]$ and $n_e = 8$ and observed that the same hyperparameters can be used to obtain good OOD detection performance in different tasks and datasets.}
\begin{table*}[hbt]
\small
\centering
\resizebox{\linewidth}{!}{
\begin{tabular}{l c   c c c  c c  c c }
\toprule
\multicolumn{1}{l}{\multirow{1}{*}{OOD}} 
& \multicolumn{6}{c}{\multirow{1}{*}{Baselines}}  & \multicolumn{2}{c}{\multirow{1}{*}{Ours (Task-Agnostic)}} \\  \cmidrule(lr){2-7}  \cmidrule(lr){8-9} 
 \multicolumn{1}{l}{Dataset} & \multicolumn{1}{c}{MSP~\cite{hendrycks2017base} } & \multicolumn{1}{c}{ODIN~\cite{liang2017enhancing} } & \multicolumn{1}{c}{Energy~\cite{liu2020energy} } & \multicolumn{1}{c}{MB~\cite{mahala18} } &  \multicolumn{1}{c}{KL~\cite{Hendrycks2019ABF} } & \multicolumn{1}{c}{Gram~\cite{sastry2021gram}} & \multicolumn{1}{c}{TAP-MB} & \multicolumn{1}{c}{TAPUDD} \\
& & & & & & & (K = 8) & (Average) 

\\ \midrule
LSUN (R) & 91.0 & 94.1 & 92.8 & 99.7 & 70.3 & 99.9 & 96.3 & 96.4 \\
LSUN (C) & 91.9 & 91.2 & 93.9 & 96.7 & 81.9 & 97.8 & 94.5 & 94.2 \\
TinyImgNet (R) & 91.0 & 94.0 & 92.4 & 99.5 & 73.8 & 99.7 & 93.8 & 94.3 \\
TinyImgNet (C) & 91.4 & 93.1 & 93.0 & 98.6 & 74.1 & 99.2 & 94.3 & 94.5 \\
SVHN & 89.9 & 96.7 & 91.2 & 99.1 & 85.5 & 99.5 & 92.8 & 93.4 \\
CIFAR100 & 86.4 & 85.8 & 87.1 & 88.2 & 69.2 & 79.0 & 88.2 & 88.9 \\
 \bottomrule
 \end{tabular}}
 \vspace{1mm}
\caption{Comparison of OOD Detection Performance of Resnet34 model trained on CIFAR10 on diverse OOD datasets measured by AUROC. The hyperparameters of ODIN and the hyperparameters and parameters of Mahalanobis are tuned using a random sample of the OOD dataset.
MB and TAP-MB refers to Mahalanobis and TAP-Mahalanobis, respectively.
}
\label{tab:ood_detection_cifar}
\end{table*}

\begin{table*}[t]
\footnotesize
\centering
\resizebox{\linewidth}{!}{
\begin{tabular}{l|c|c c c c c c c c c c | c}
\toprule
\multirow{1}{*}{\textbf{Method}} & \multicolumn{1}{c|}{\textbf{Network}} & \multicolumn{1}{c}{\textbf{Airplane}}    & \multicolumn{1}{c}{\textbf{Automobile}}            & \multicolumn{1}{c}{\textbf{Bird}}         & \multicolumn{1}{c}{\textbf{Cat}}       & \multicolumn{1}{c}{\textbf{Deer}}   & \multicolumn{1}{c}{\textbf{Dog}}    & \multicolumn{1}{c}{\textbf{Frog}}            & \multicolumn{1}{c}{\textbf{Horse}}         & \multicolumn{1}{c}{\textbf{Ship}}       & \multicolumn{1}{c|}{\textbf{Truck}}  & \multicolumn{1}{c}{\textbf{Average}}     \\  
\midrule
VAE & - & 70.4 & 38.6 & 67.9 & 53.5 & 74.8 & 52.3 & 68.7 & 49.3 & 69.6 & 38.6 & 58.4 \\
OCSVM & - & 63.0 & 44.0 & 64.9 & 48.7 & 73.5 & 50.0 & 72.5 & 53.3 & 64.9 & 50.8 & 58.6 \\
AnoGAN & DCGAN & 67.1 & 54.7 & 52.9 & 54.5 & 65.1 & 60.3 & 58.5 & 62.5 & 75.8 & 66.5 & 61.8 \\
PixelCNN & PixelCNN & 53.1 & 99.5 & 47.6 & 51.7 & 73.9 & 54.2 & 59.2 & 78.5 & 34.0 & 66.2 & 61.8 \\
DSVDD & LeNet & 61.7 & 65.9 & 50.8 & 59.1 & 60.9 & 65.7 & 67.7 & 67.3 & 75.9 & 73.1 & 64.8 \\
OCGAN & OCGAN & 75.7 & 53.1 & 64.0 & 62.0 & 72.3 & 62.0 & 72.3 & 57.5 & 82.0 & 55.4 & 65.6 \\
Reconstruction Error & Resnet18 & 71.5 & 39.2 & 68.7 & 55.9 & 72.6 & 54.4 & 63.3 & 49.1 & 71.3 & 37.6 & 58.4 \\
MB (K = 1) & Resnet18 & 69.2 & 66.9 & 66.3 & 52.3 & 74.8 & 50.7 & 77.8 & 52.9 & 64.0 & 52.1 & 62.7 \\ \midrule
TAP-MB (K = 1) & Resnet18 & 69.2 & 66.8 & 66.2 & 52.3 & 74.8 & 50.7 & 77.8 & 52.9 & 64.0 & 52.1 & 62.7 \\
TAP-MB (K = 2) & Resnet18 & 73.0 & 67.3 & 66.5 & 50.7 & 74.4 & 49.9 & 74.7 & 52.4 & 66.2 & 52.6 & 62.8 \\
TAP-MB (K = 3) & Resnet18 & 73.4 & 68.0 & 66.9 & 54.9 & 74.4 & 50.2 & 76.1 & 51.7 & 61.3 & 52.3 & \textbf{62.9} \\
TAP-MB (K = 4) & Resnet18 & 68.1 & 64.1 & 63.6 & 56.6 & 73.5 & 48.7 & 75.3 & 49.5 & 54.8 & 48.6 & 60.3 \\
\bottomrule
\end{tabular}
}
\vspace{1mm}
\renewcommand{\arraystretch}{1.1}
\caption{Comparison of TAP-Mahalanobis with other detectors for anomaly detection task on CIFAR-10 dataset. TAP-MB and MB denotes TAP-Mahalanobis and Mahalanobis, respectively.
}
\label{table:anomaly detection}
\end{table*}

\section{Experimental Details}
\label{sec:supp_experimental_details}

For binary classification and regression tasks, we use the RSNA Bone Age dataset~\cite{013a4b4eeca44ecab48a66956acbb91d}, a real-world dataset that contains $12611$ left-hand X-ray images of the patient, along with their gender and age ($0$ to $20$ years). We randomly split the dataset in $8$:$1$:$1$ ratio to form train, val, and test split with $9811$, $1400$, and $1400$ samples, respectively.
Following~\cite{Park2021NaturalAS}, to reflect diverse X-ray imaging set-ups in the hospital, 
we vary the brightness factor of the test set between $0$ and $6.5$ and form $20$ different NAS datasets.
In-distribution data comprises images with a brightness factor of $1.0$ (unmodified images).

\parahead{Binary Classification}
 We use a ResNet18~\cite{He2016DeepRL} model, pretrained on ImageNet~\cite{Deng2009ImageNetAL}, and add two fully-connected layers containing $128$ and $2$ hidden units with a relu activation. We train the network to classify gender given the x-ray image.
Each model is trained for $100$ epochs using SGD optimizer with a learning rate of $0.001$ and momentum of $0.9$, using a batch size of $64$.

\parahead{Regression}
 We use a ResNet18~\cite{He2016DeepRL} model that is pretrained on ImageNet~\cite{Deng2009ImageNetAL} and train it to predict the age given the x-ray image.  After the average pooling layer, we add two fully-connected layers with $128$ and $1$ units with a relu activation.
Each model is trained for $100$ epochs using SGD optimizer with a learning rate of $1e-05$, weight decay of $0.0001$, and momentum of $0.9$, using a batch size of $64$. We also apply gradient clipping with a clip value of $5.0$.

\vspace{-0.9cm}
The results are measured by computing mean and standard deviation across $10$ trials upon randomly chosen seeds. We perform all experiments on NVIDIA GeForce RTX A6000 GPUs.
\section{Additional Results}
\label{sec:supp_additional_exp}

\subsection{Evaluation on Synthetic Datasets}

\parahead{TAPUDD outperforms TAP-Mahalanobis}
We  present a comparison of TAPUDD against TAP-Mahalanobis on 2-D synthetic dataset for binary classification task, in continuation to the discussion in Section 4.1. \cref{fig:synthetic_data_umahala_vs_tauood_binary} presents the ID score landscape of TAP-Mahalanobis for different values of $K$ and TAPUDD with different ensemble variations for binary classification in a 2-D synthetic dataset.
The \textbf{\textcolor{magenta}{Pink Points}} represent the in-distribution data. We observe that for certain values of $K$,  TAP-Mahalanobis fails to detect some OOD samples. However, all ensemble variations of TAPUDD effectively detect OOD samples and performs better, or on par, with TAP-Mahalanobis. 
Thus, TAPUDD eliminates the necessity of choosing the optimal value of $K$.

\subsection{OOD Detection in Multi-class Classification}
\label{sec:cifar_level}
As stated in Section 4 of the main paper, we also evaluate our approach for OOD detection in multi-class classification task on benchmark datasets to further bolster the effectiveness of our proposed approach. 
We use the pretrained ResNet34~\cite{He2016DeepRL} model trained on CIFAR-10 dataset (opensourced in \cite{mahala18}).  We consider the test set as the in-distribution samples and evaluate our approach on diverse OOD datasets used in literature (TinyImagenet, LSUN~\cite{Yu2015LSUNCO}, SVHN~\cite{Netzer2011ReadingDI} and CIFAR100). \cref{tab:ood_detection_cifar} presents the OOD detection performance of our approach and baselines based on AUROC score. We observe that our task-agnostic and post-hoc approach performs better or comparable to the baselines.

\vspace{-1mm}
\subsection{Anomaly Detection}
\label{sec:anomaly_detection}

We evaluate our approach for anomaly detection task in which one of the CIFAR-10 classes is considered as in-distribution and the samples from rest of the classes are considered as anomalous. We train a Resnet18 based auto-encoder model using MSE loss function which aims to minimize the reconstruction error between the output and input image. 
Since, in this work, we provide a task-agnostic and post-hoc approach to detect samples from unseen distribution, we consider that we are given a model trained on one of the classes CIFAR10 and our objective is to detect anomalous samples (\ie , samples from other classes of CIFAR10). 
We first compare our approach with two baselines that does not modify the base architecture.
Reconstruction-error based baseline which rely on reconstruction error to determine if a sample is anomalous or not. Mahalanobis distance based detector with number of classes as $1$ to detect anomalous samples.
Further, we also compare our approach with various well-known baselines for anomaly detection, including VAE, OCSVM, AnoGAN, PixelCNN, DSVDD, OCGAN.
Although it is unfair to compare with these baselines as they have different base models, we compare against these baselines since they are used widely in literature.
We do not compare with other baselines including CSI, SSD as they might fail in certain scenarios (described in \cref{sec:supp_analysis}). ~\cref{table:anomaly detection} presents a comparison of TAP-Mahalanobis with other detectors for anomaly detection task on CIFAR-10 dataset. We observe that our task-agnostic and post-hoc approach is better than reconstruction-error based baseline. Our approach is also better than or on-par with other baselines used in the literature. This demonstrates the effectiveness of our approach on anomaly detection task. 

\section{Quantitative Results with Different Performance Metrics}
\label{sec:supp_other_metrics}
\begin{table}[h]
\vspace{1mm}
\footnotesize
\centering
\resizebox{\columnwidth}{!}{
\begin{tabular}{l|c|c|c|c|c}
\toprule
\multirow{1}{*}{\textbf{Method}} & \multicolumn{1}{c|}{\textbf{iNaturalist}}    & \multicolumn{1}{c|}{\textbf{SUN}}            & \multicolumn{1}{c|}{\textbf{Places}}         & \multicolumn{1}{c|}{\textbf{Textures}}       & \multicolumn{1}{c}{\textbf{Average}}        \\ \midrule
Expected & Low & Low & Low & \textbf{High} & -- \\
\midrule
MSP~\cite{hendrycks2017base} & 97.26 & 94.41 & 94.12 & 95.65 & 95.36 \\
ODIN~\cite{liang2017enhancing} & 97.80 & 96.23 & 95.33 & 96.11 & 96.37 \\
\rowcolor{teagreen} Mahalanobis \cite{mahala18} & 87.35 & 90.32 & 90.25 & 92.52 & 90.11 \\
Energy~\cite{liu2020energy} & 97.62 & 96.55 & 95.47 & 96.04 & 96.42 \\
KL Matching~\cite{Hendrycks2019ABF} & 97.98& 94.11 & 93.62 & 97.96 & 95.92 \\
MOS~\cite{Huang2021MOSTS} & 99.62 & 98.17 & 97.36 & 96.68 & 97.96 \\ \midrule
\rowcolor{teagreen} TAPUDD (Average) & \textbf{91.87} & \textbf{91.02} & 90.08 & \textbf{99.68} & \textbf{93.16} \\
\bottomrule
\end{tabular}
}
\vspace{1mm}
\caption{OOD detection performance comparison between TAPUDD method and baselines measured by AUPR. 
\edit{
Ideally, all methods should follow the expected results obtained from our analysis (described in first row in \textcolor{ForestGreen}{green} color) conducted in Section 4.4 of the main paper.
However, as highlighted in \textcolor{ForestGreen}{green} color, only Mahalanobis and our proposed approach follow the expected results. This highlights the failure of existing baselines, including MSP, ODIN, Energy, KL Matching, and MOS.
Further, amongst all methods following the expected results (highlighted in \textcolor{ForestGreen}{green} color), \textit{our approach is highly sensitive to OOD samples and significantly outperforms the baselines}. }
}
\label{table:multi_class_ood_aupr}
\end{table}
We report additional metrics to evaluate the unseen distribution detection performance of baselines and our approach in binary classification, regression, and large-scale classification tasks.
In \cref{tab:binary_class_nas_aupr} and \cref{tab:binary_class_nas_fpr}, we compare the NAS detection performance of baselines and our approach in binary classification task based on AUPR and FPR95, respectively. 
We also report the NAS detection performance of baselines and our method in regression task based on AUPR and FPR95 in \cref{tab:regression_nas_aupr} and \cref{tab:regression_fpr95}, respectively.
Results demonstrate that our proposed approaches, TAPUDD and TAP-Mahalanobis are more sensitive to NAS samples compared to competitive baselines.
Further, we report AUPR to evaluate the OOD detection performance of different methods in large-scale classification task \cref{table:multi_class_ood_aupr}.
As expected from the analysis conducted in Section 4.4 of the main paper, the results indicate that our approach detects samples from the Textures dataset as more OOD compared to samples from iNaturalist, SUN, and Places (similar to the way humans perceive).

\begin{table*}[t]
\small
\centering
\resizebox{0.98\linewidth}{!}{
\begin{tabular}{c  c  c  c c c  c c c c c }
\toprule
\multicolumn{1}{c}{\multirow{1}{*}{Brightness}} 
& \multicolumn{7}{c}{\multirow{1}{*}{Baselines}}  & \multicolumn{3}{c}{\multirow{1}{*}{Ours (Task-Agnostic)}} \\  \cmidrule(lr){2-8}  \cmidrule(lr){9-11} 
\multicolumn{1}{c}{} & \multicolumn{1}{c}{MSP~\cite{hendrycks2017base} } & \multicolumn{1}{c}{ODIN~\cite{liang2017enhancing} } & \multicolumn{1}{c}{Energy~\cite{liu2020energy} } & \multicolumn{1}{c}{MB~\cite{mahala18} } &  \multicolumn{1}{c}{KL~\cite{Hendrycks2019ABF} } &  \multicolumn{1}{c}{MOS~\cite{Huang2021MOSTS} } & \multicolumn{1}{c}{Gram~\cite{sastry2021gram}}  & \multicolumn{1}{c}{TAP-MOS} & \multicolumn{1}{c}{TAP-MB} & \multicolumn{1}{c}{TAPUDD} \\
& & & & & & (K = 8) & & (K = 8) & (K = 8) & (Average) 

\\ \midrule
        
0.0 & 94.1{\scriptsize $\pm$2.6} & 94.1{\scriptsize $\pm$2.6} & 93.9{\scriptsize $\pm$2.8} & 100.0{\scriptsize $\pm$0.0} & 51.4{\scriptsize $\pm$23.3} & 94.5{\scriptsize $\pm$2.9} & 99.7{\scriptsize $\pm$0.7} & 63.1{\scriptsize $\pm$14.3} & 100.0{\scriptsize $\pm$0.1} & 100.0{\scriptsize $\pm$0.0} \\
0.2 & 68.2{\scriptsize $\pm$4.3} & 68.0{\scriptsize $\pm$4.5} & 67.8{\scriptsize $\pm$4.8} & 89.0{\scriptsize $\pm$4.6} & 44.9{\scriptsize $\pm$1.8} & 67.5{\scriptsize $\pm$4.5} & 73.9{\scriptsize $\pm$1.5} & 67.4{\scriptsize $\pm$9.0} & 89.3{\scriptsize $\pm$3.5} & 89.7{\scriptsize $\pm$4.0} \\
0.4 & 57.6{\scriptsize $\pm$2.2} & 56.6{\scriptsize $\pm$2.3} & 56.4{\scriptsize $\pm$2.6} & 71.7{\scriptsize $\pm$4.0} & 47.4{\scriptsize $\pm$0.7} & 56.7{\scriptsize $\pm$1.8} & 71.0{\scriptsize $\pm$1.2} & 57.5{\scriptsize $\pm$4.3} & 72.6{\scriptsize $\pm$3.7} & 73.2{\scriptsize $\pm$4.4} \\
0.6 & 53.9{\scriptsize $\pm$2.0} & 52.2{\scriptsize $\pm$1.4} & 52.1{\scriptsize $\pm$1.3} & 59.9{\scriptsize $\pm$2.7} & 48.8{\scriptsize $\pm$0.7} & 52.5{\scriptsize $\pm$1.2} & 70.3{\scriptsize $\pm$1.4} & 53.2{\scriptsize $\pm$2.3} & 60.8{\scriptsize $\pm$2.6} & 60.9{\scriptsize $\pm$2.9} \\
0.8 & 52.3{\scriptsize $\pm$1.7} & 50.2{\scriptsize $\pm$0.7} & 50.2{\scriptsize $\pm$0.7} & 52.4{\scriptsize $\pm$1.3} & 49.6{\scriptsize $\pm$0.4} & 50.4{\scriptsize $\pm$0.6} & 70.0{\scriptsize $\pm$1.4} & 50.7{\scriptsize $\pm$1.3} & 52.5{\scriptsize $\pm$1.4} & 52.5{\scriptsize $\pm$1.5} \\
\rowcolor{Gray} 1.0 & 52.2{\scriptsize $\pm$1.4} & 50.0{\scriptsize $\pm$0.0} & 50.0{\scriptsize $\pm$0.0} & 50.0{\scriptsize $\pm$0.0} & 50.0{\scriptsize $\pm$0.0} & 50.0{\scriptsize $\pm$0.0} & 69.9{\scriptsize $\pm$1.4} & 50.0{\scriptsize $\pm$0.0} & 50.0{\scriptsize $\pm$0.0} & 50.0{\scriptsize $\pm$0.0} \\
1.2 & 53.4{\scriptsize $\pm$1.0} & 51.4{\scriptsize $\pm$0.7} & 51.5{\scriptsize $\pm$0.7} & 55.0{\scriptsize $\pm$1.8} & 48.9{\scriptsize $\pm$0.5} & 51.4{\scriptsize $\pm$0.7} & 70.2{\scriptsize $\pm$1.4} & 50.9{\scriptsize $\pm$0.4} & 56.2{\scriptsize $\pm$1.6} & 56.1{\scriptsize $\pm$1.6} \\
1.4 & 56.6{\scriptsize $\pm$0.9} & 55.2{\scriptsize $\pm$1.3} & 55.2{\scriptsize $\pm$1.3} & 62.8{\scriptsize $\pm$2.6} & 47.7{\scriptsize $\pm$0.7} & 55.1{\scriptsize $\pm$1.2} & 71.0{\scriptsize $\pm$1.3} & 53.0{\scriptsize $\pm$0.8} & 64.2{\scriptsize $\pm$2.2} & 64.0{\scriptsize $\pm$2.2} \\
1.6 & 60.4{\scriptsize $\pm$1.6} & 59.3{\scriptsize $\pm$2.2} & 59.4{\scriptsize $\pm$2.3} & 70.8{\scriptsize $\pm$3.1} & 46.9{\scriptsize $\pm$0.9} & 59.0{\scriptsize $\pm$1.8} & 71.8{\scriptsize $\pm$1.2} & 55.1{\scriptsize $\pm$1.5} & 72.2{\scriptsize $\pm$2.5} & 72.0{\scriptsize $\pm$2.6} \\
1.8 & 63.9{\scriptsize $\pm$2.8} & 63.2{\scriptsize $\pm$3.4} & 63.4{\scriptsize $\pm$3.6} & 77.6{\scriptsize $\pm$3.2} & 46.9{\scriptsize $\pm$1.3} & 62.5{\scriptsize $\pm$2.9} & 72.7{\scriptsize $\pm$1.2} & 56.5{\scriptsize $\pm$3.5} & 78.5{\scriptsize $\pm$2.8} & 78.3{\scriptsize $\pm$3.0} \\
2.0 & 66.5{\scriptsize $\pm$4.4} & 66.0{\scriptsize $\pm$5.1} & 66.3{\scriptsize $\pm$5.2} & 83.0{\scriptsize $\pm$3.0} & 47.4{\scriptsize $\pm$1.6} & 65.0{\scriptsize $\pm$4.5} & 73.6{\scriptsize $\pm$1.3} & 57.6{\scriptsize $\pm$5.4} & 83.6{\scriptsize $\pm$2.7} & 83.3{\scriptsize $\pm$2.8} \\
2.5 & 71.0{\scriptsize $\pm$7.5} & 70.5{\scriptsize $\pm$8.4} & 70.7{\scriptsize $\pm$8.5} & 91.6{\scriptsize $\pm$2.5} & 48.4{\scriptsize $\pm$3.4} & 69.4{\scriptsize $\pm$7.2} & 75.7{\scriptsize $\pm$2.2} & 60.3{\scriptsize $\pm$7.6} & 91.3{\scriptsize $\pm$2.9} & 91.1{\scriptsize $\pm$3.0} \\
3.0 & 75.1{\scriptsize $\pm$9.5} & 74.7{\scriptsize $\pm$10.3} & 74.9{\scriptsize $\pm$10.4} & 95.7{\scriptsize $\pm$1.6} & 48.5{\scriptsize $\pm$4.5} & 73.9{\scriptsize $\pm$9.2} & 77.8{\scriptsize $\pm$3.2} & 63.8{\scriptsize $\pm$8.1} & 95.0{\scriptsize $\pm$2.8} & 94.9{\scriptsize $\pm$2.9} \\
3.5 & 76.5{\scriptsize $\pm$10.2} & 76.3{\scriptsize $\pm$10.7} & 76.4{\scriptsize $\pm$10.8} & 97.5{\scriptsize $\pm$1.1} & 49.4{\scriptsize $\pm$5.5} & 75.4{\scriptsize $\pm$9.7} & 79.2{\scriptsize $\pm$4.0} & 65.9{\scriptsize $\pm$8.9} & 96.5{\scriptsize $\pm$2.7} & 96.5{\scriptsize $\pm$2.7} \\
4.0 & 78.8{\scriptsize $\pm$9.9} & 78.8{\scriptsize $\pm$10.0} & 78.7{\scriptsize $\pm$10.4} & 98.4{\scriptsize $\pm$0.6} & 48.6{\scriptsize $\pm$6.2} & 78.1{\scriptsize $\pm$9.3} & 80.9{\scriptsize $\pm$5.0} & 67.9{\scriptsize $\pm$9.4} & 97.4{\scriptsize $\pm$2.5} & 97.3{\scriptsize $\pm$2.4} \\
4.5 & 81.0{\scriptsize $\pm$8.3} & 81.1{\scriptsize $\pm$8.3} & 81.0{\scriptsize $\pm$8.8} & 98.9{\scriptsize $\pm$0.4} & 46.5{\scriptsize $\pm$6.4} & 80.6{\scriptsize $\pm$7.9} & 82.5{\scriptsize $\pm$4.8} & 69.5{\scriptsize $\pm$9.5} & 98.0{\scriptsize $\pm$2.1} & 98.0{\scriptsize $\pm$1.8} \\
5.0 & 82.9{\scriptsize $\pm$6.6} & 82.9{\scriptsize $\pm$6.6} & 82.8{\scriptsize $\pm$7.2} & 99.1{\scriptsize $\pm$0.3} & 45.0{\scriptsize $\pm$6.0} & 82.7{\scriptsize $\pm$6.3} & 84.1{\scriptsize $\pm$4.7} & 70.3{\scriptsize $\pm$9.8} & 98.4{\scriptsize $\pm$1.8} & 98.5{\scriptsize $\pm$1.3} \\
5.5 & 84.4{\scriptsize $\pm$5.4} & 84.4{\scriptsize $\pm$5.5} & 84.3{\scriptsize $\pm$6.1} & 99.3{\scriptsize $\pm$0.3} & 44.5{\scriptsize $\pm$6.0} & 84.3{\scriptsize $\pm$5.3} & 85.5{\scriptsize $\pm$4.8} & 70.7{\scriptsize $\pm$10.3} & 98.7{\scriptsize $\pm$1.6} & 98.8{\scriptsize $\pm$1.0} \\
6.0 & 85.7{\scriptsize $\pm$4.7} & 85.7{\scriptsize $\pm$4.7} & 85.6{\scriptsize $\pm$5.3} & 99.4{\scriptsize $\pm$0.3} & 44.2{\scriptsize $\pm$5.8} & 85.5{\scriptsize $\pm$4.6} & 86.6{\scriptsize $\pm$4.8} & 71.1{\scriptsize $\pm$10.9} & 98.9{\scriptsize $\pm$1.4} & 99.0{\scriptsize $\pm$0.7} \\
6.5 & 86.5{\scriptsize $\pm$4.3} & 86.5{\scriptsize $\pm$4.3} & 86.4{\scriptsize $\pm$4.8} & 99.4{\scriptsize $\pm$0.3} & 44.4{\scriptsize $\pm$5.5} & 86.3{\scriptsize $\pm$4.2} & 87.5{\scriptsize $\pm$4.8} & 71.2{\scriptsize $\pm$11.4} & 99.0{\scriptsize $\pm$1.2} & 99.1{\scriptsize $\pm$0.6} \\
\midrule
Average & 70.1 & 69.4 & 69.4 & 82.6 & 47.5 & 69.0 & 77.7 & 62.0 & 81.7 & \textbf{81.9} \\
 \bottomrule
 \end{tabular}}
 \vspace{1mm}
\caption{NAS detection performance in binary classification task (gender prediction) for NAS shift of brightness in RSNA boneage dataset measured by AUPR. Highlighted row presents the performance on in-distribution dataset. MB and TAP-MB refers to Mahalanobis and TAP-Mahalanobis, respectively.
}
\label{tab:binary_class_nas_aupr}
\vspace{2mm}
\end{table*}
        
\begin{table*}[t]
\small
\centering
\resizebox{0.98\linewidth}{!}{
\begin{tabular}{c  c  c  c c c  c c c c c }
\toprule
\multicolumn{1}{c}{\multirow{1}{*}{Brightness}} 
& \multicolumn{7}{c}{\multirow{1}{*}{Baselines}}  & \multicolumn{3}{c}{\multirow{1}{*}{Ours (Task-Agnostic)}} \\  \cmidrule(lr){2-8}  \cmidrule(lr){9-11} 
\multicolumn{1}{c}{} & \multicolumn{1}{c}{MSP~\cite{hendrycks2017base} } & \multicolumn{1}{c}{ODIN~\cite{liang2017enhancing} } & \multicolumn{1}{c}{Energy~\cite{liu2020energy} } & \multicolumn{1}{c}{MB~\cite{mahala18} } &  \multicolumn{1}{c}{KL~\cite{Hendrycks2019ABF} } &  \multicolumn{1}{c}{MOS~\cite{Huang2021MOSTS} } & \multicolumn{1}{c}{Gram~\cite{sastry2021gram}}  & \multicolumn{1}{c}{TAP-MOS} & \multicolumn{1}{c}{TAP-MB} & \multicolumn{1}{c}{TAPUDD} \\
& & & & & & (K = 8) & & (K = 8) & (K = 8) & (Average) 

\\ \midrule
        
0.0 & 90.0{\scriptsize $\pm$31.6} & 90.0{\scriptsize $\pm$31.6} & 90.0{\scriptsize $\pm$31.6} & 0.0{\scriptsize $\pm$0.0} & 90{\scriptsize $\pm$31.6} & 80{\scriptsize $\pm$42.2} & 0.0{\scriptsize $\pm$0.0} & 100.0{\scriptsize $\pm$0.0} & 0{.0\scriptsize $\pm$0.0} & 0.0{\scriptsize $\pm$0.0} \\
0.2 & 90.4{\scriptsize $\pm$1.6} & 90.4{\scriptsize $\pm$1.6} & 90.6{\scriptsize $\pm$2.5} & 61.4{\scriptsize $\pm$14.8} & 92.1{\scriptsize $\pm$1.3} & 90.7{\scriptsize $\pm$1.5} & 88.5{\scriptsize $\pm$2.7} & 86.0{\scriptsize $\pm$9.2} & 65.6{\scriptsize $\pm$11.9} & 62.5{\scriptsize $\pm$15.1} \\
0.4 & 93.9{\scriptsize $\pm$1.0} & 93.9{\scriptsize $\pm$1.0} & 93.9{\scriptsize $\pm$1.1} & 86.4{\scriptsize $\pm$5.9} & 94.5{\scriptsize $\pm$1.0} & 94.0{\scriptsize $\pm$0.9} & 93.6{\scriptsize $\pm$0.9} & 91.6{\scriptsize $\pm$5.0} & 89.1{\scriptsize $\pm$3.3} & 88.5{\scriptsize $\pm$4.2} \\
0.6 & 94.3{\scriptsize $\pm$0.8} & 94.3{\scriptsize $\pm$0.8} & 94.3{\scriptsize $\pm$0.7} & 92.8{\scriptsize $\pm$2.0} & 94.5{\scriptsize $\pm$1.1} & 94.4{\scriptsize $\pm$1.0} & 94.2{\scriptsize $\pm$1.2} & 93.4{\scriptsize $\pm$2.6} & 93.3{\scriptsize $\pm$1.3} & 93.4{\scriptsize $\pm$1.2} \\
0.8 & 95.0{\scriptsize $\pm$0.4} & 95.0{\scriptsize $\pm$0.4} & 95.0{\scriptsize $\pm$0.4} & 94.7{\scriptsize $\pm$0.8} & 95.5{\scriptsize $\pm$0.7} & 95.2{\scriptsize $\pm$0.5} & 94.9{\scriptsize $\pm$0.6} & 94.8{\scriptsize $\pm$1.2} & 95.4{\scriptsize $\pm$0.7} & 95.2{\scriptsize $\pm$0.5} \\
\rowcolor{Gray} 1.0 & 95.0{\scriptsize $\pm$0.0} & 95.0{\scriptsize $\pm$0.0} & 95.0{\scriptsize $\pm$0.0} & 95.0{\scriptsize $\pm$0.0} & 95.0{\scriptsize $\pm$0.0} & 95.0{\scriptsize $\pm$0.0} & 95.0{\scriptsize $\pm$0.0} & 95.0{\scriptsize $\pm$0.0} & 95.0{\scriptsize $\pm$0.0} & 95.0{\scriptsize $\pm$0.0} \\
1.2 & 94.8{\scriptsize $\pm$0.3} & 94.8{\scriptsize $\pm$0.3} & 94.5{\scriptsize $\pm$0.3} & 93.5{\scriptsize $\pm$1.0} & 94.7{\scriptsize $\pm$1.3} & 94.8{\scriptsize $\pm$0.4} & 94.6{\scriptsize $\pm$0.7} & 95.0{\scriptsize $\pm$0.8} & 93.5{\scriptsize $\pm$0.9} & 93.3{\scriptsize $\pm$0.6} \\
1.4 & 93.3{\scriptsize $\pm$0.7} & 93.3{\scriptsize $\pm$0.7} & 93.0{\scriptsize $\pm$0.8} & 90.0{\scriptsize $\pm$1.8} & 93.5{\scriptsize $\pm$1.6} & 93.6{\scriptsize $\pm$0.9} & 93.0{\scriptsize $\pm$0.8} & 94.4{\scriptsize $\pm$1.9} & 89.9{\scriptsize $\pm$2.2} & 89.1{\scriptsize $\pm$2.1} \\
1.6 & 92.2{\scriptsize $\pm$0.6} & 92.2{\scriptsize $\pm$0.6} & 91.8{\scriptsize $\pm$0.8} & 83.5{\scriptsize $\pm$2.3} & 93.1{\scriptsize $\pm$0.9} & 92.1{\scriptsize $\pm$1} & 90.8{\scriptsize $\pm$1.6} & 93.4{\scriptsize $\pm$2.6} & 84.1{\scriptsize $\pm$3.9} & 83.4{\scriptsize $\pm$3.7} \\
1.8 & 91.1{\scriptsize $\pm$1.2} & 91.1{\scriptsize $\pm$1.2} & 90.5{\scriptsize $\pm$1.3} & 77.1{\scriptsize $\pm$3.3} & 92.1{\scriptsize $\pm$1.1} & 91.0{\scriptsize $\pm$1.4} & 88.5{\scriptsize $\pm$2.4} & 92.5{\scriptsize $\pm$3.6} & 77.3{\scriptsize $\pm$6.1} & 76.8{\scriptsize $\pm$5.4} \\
2.0 & 90.2{\scriptsize $\pm$1.1} & 90.2{\scriptsize $\pm$1.1} & 89.5{\scriptsize $\pm$1.2} & 70.5{\scriptsize $\pm$5.6} & 91.6{\scriptsize $\pm$1.4} & 90.3{\scriptsize $\pm$1.2} & 86.6{\scriptsize $\pm$3.4} & 92.5{\scriptsize $\pm$4.3} & 70.1{\scriptsize $\pm$8.7} & 69.9{\scriptsize $\pm$8.4} \\
2.5 & 87.4{\scriptsize $\pm$2.5} & 87.4{\scriptsize $\pm$2.5} & 87.0{\scriptsize $\pm$2.6} & 51.6{\scriptsize $\pm$11.3} & 89.8{\scriptsize $\pm$2.1} & 87.7{\scriptsize $\pm$2.5} & 80.5{\scriptsize $\pm$5.3} & 91.5{\scriptsize $\pm$5.6} & 49.8{\scriptsize $\pm$16.1} & 50.5{\scriptsize $\pm$15.0} \\
3.0 & 86.2{\scriptsize $\pm$4.9} & 86.2{\scriptsize $\pm$4.9} & 85.7{\scriptsize $\pm$4.6} & 32.5{\scriptsize $\pm$12.4} & 88.8{\scriptsize $\pm$3.9} & 86.5{\scriptsize $\pm$4.9} & 76.0{\scriptsize $\pm$7.9} & 89.6{\scriptsize $\pm$8.1} & 33.5{\scriptsize $\pm$21.1} & 34.3{\scriptsize $\pm$18.7} \\
3.5 & 84.6{\scriptsize $\pm$6.9} & 84.7{\scriptsize $\pm$6.9} & 84.7{\scriptsize $\pm$6.3} & 19.0{\scriptsize $\pm$10.2} & 87.8{\scriptsize $\pm$5.7} & 85.5{\scriptsize $\pm$6.7} & 72.4{\scriptsize $\pm$11.9} & 87.9{\scriptsize $\pm$12.3} & 25.4{\scriptsize $\pm$23.7} & 25.3{\scriptsize $\pm$20.9} \\
4.0 & 83.6{\scriptsize $\pm$8.2} & 83.6{\scriptsize $\pm$8.1} & 83.5{\scriptsize $\pm$7.4} & 10.8{\scriptsize $\pm$6.9} & 86.5{\scriptsize $\pm$6.8} & 84.1{\scriptsize $\pm$8.1} & 68.2{\scriptsize $\pm$14.3} & 85.1{\scriptsize $\pm$16} & 19.8{\scriptsize $\pm$24.5} & 19.0{\scriptsize $\pm$20.9} \\
4.5 & 81.9{\scriptsize $\pm$7.4} & 81.9{\scriptsize $\pm$7.4} & 81.7{\scriptsize $\pm$7.6} & 5.4{\scriptsize $\pm$3.6} & 84.9{\scriptsize $\pm$6.4} & 82.3{\scriptsize $\pm$7.6} & 64.7{\scriptsize $\pm$13.5} & 83.6{\scriptsize $\pm$16.8} & 14.8{\scriptsize $\pm$22.4} & 14.1{\scriptsize $\pm$18.1} \\
5.0 & 80.5{\scriptsize $\pm$7.4} & 80.5{\scriptsize $\pm$7.3} & 80.0{\scriptsize $\pm$7.8} & 2.7{\scriptsize $\pm$1.9} & 83.9{\scriptsize $\pm$6.2} & 81.2{\scriptsize $\pm$7.2} & 61.7{\scriptsize $\pm$13.4} & 82.7{\scriptsize $\pm$15.9} & 11.4{\scriptsize $\pm$19.9} & 10.0{\scriptsize $\pm$13.9} \\
5.5 & 78.6{\scriptsize $\pm$8.1} & 78.6{\scriptsize $\pm$8.1} & 78.2{\scriptsize $\pm$8.6} & 1.8{\scriptsize $\pm$1.7} & 82.3{\scriptsize $\pm$6.9} & 79.3{\scriptsize $\pm$8.0} & 58.3{\scriptsize $\pm$13.9} & 82.8{\scriptsize $\pm$14.8} & 9.2{\scriptsize $\pm$18.0} & 7.1{\scriptsize $\pm$10.4} \\
6.0 & 77.6{\scriptsize $\pm$8.9} & 77.6{\scriptsize $\pm$8.9} & 77.0{\scriptsize $\pm$9.4} & 1.3{\scriptsize $\pm$1.6} & 81.7{\scriptsize $\pm$7.7} & 78.6{\scriptsize $\pm$8.8} & 56.1{\scriptsize $\pm$14.6} & 83.7{\scriptsize $\pm$14} & 7.5{\scriptsize $\pm$16.3} & 5.1{\scriptsize $\pm$7.6} \\
6.5 & 76.5{\scriptsize $\pm$9.7} & 76.6{\scriptsize $\pm$9.6} & 76.1{\scriptsize $\pm$10.3} & 1.0{\scriptsize $\pm$1.5} & 80.3{\scriptsize $\pm$8} & 77.7{\scriptsize $\pm$9.2} & 54.2{\scriptsize $\pm$15.5} & 84.2{\scriptsize $\pm$13.5} & 6.3{\scriptsize $\pm$14.9} & 3.6{\scriptsize $\pm$5.9} \\
\midrule
Average & 87.9	& 87.9	& 87.6	& 48.6	& 89.6	& 87.7	& 75.6	& 62.0	& 81.7	& \textbf{46.5} \\
 \bottomrule
 \end{tabular}}
 \vspace{1mm}
\caption{NAS detection performance in binary classification task (gender prediction) for NAS shift of brightness in RSNA boneage dataset measured by FPR95. Highlighted row presents the performance on in-distribution dataset. MB and TAP-MB refers to Mahalanobis and TAP-Mahalanobis, respectively.
}
\label{tab:binary_class_nas_fpr}
\end{table*}
        
\begin{table*}[hbt]
\footnotesize
\centering
\resizebox{0.78\linewidth}{!}{
\begin{tabular}{c  c  c  c c c  c }
\toprule
\multicolumn{1}{c}{\multirow{1}{*}{Brightness}} 
& \multicolumn{3}{c}{\multirow{1}{*}{Baselines}}  & \multicolumn{3}{c}{\multirow{1}{*}{Ours (Task-Agnostic)}} \\  \cmidrule(lr){2-4}  \cmidrule(lr){5-7} 
\multicolumn{1}{c}{} & \multicolumn{1}{c}{DE~\cite{lakshminarayanan2017simple} } & \multicolumn{1}{c}{MC Dropout~\cite{gal2016dropout} } & \multicolumn{1}{c}{$\text{SWAG}^{*}$~\cite{maddox_2019_simple}}  & \multicolumn{1}{c}{TAP-MOS} & \multicolumn{1}{c}{TAP-MB} & \multicolumn{1}{c}{TAPUDD}\\
& & & & (K = 8) & (K = 8) & (Average) 

\\ \midrule
0.0 & 100.0{\scriptsize $\pm$NA} & 34.9{\scriptsize $\pm$NA} & 100.0{\scriptsize $\pm$NA} & 74.4{\scriptsize $\pm$20.8} & 99.8{\scriptsize $\pm$0.4} & 100.0{\scriptsize $\pm$0.0} \\
0.2 & 53.9{\scriptsize $\pm$NA} & 48.4{\scriptsize $\pm$NA} & 51.4{\scriptsize $\pm$NA} & 69.2{\scriptsize $\pm$17.2} & 89.6{\scriptsize $\pm$12.8} & 89.6{\scriptsize $\pm$6.1} \\
0.4 & 50.0{\scriptsize $\pm$NA} & 51.0{\scriptsize $\pm$NA} & 49.4{\scriptsize $\pm$NA} & 69.2{\scriptsize $\pm$16.9} & 75.6{\scriptsize $\pm$15.8} & 68.1{\scriptsize $\pm$4.8} \\
0.6 & 50.1{\scriptsize $\pm$NA} & 50.4{\scriptsize $\pm$NA} & 49.2{\scriptsize $\pm$NA} & 64.3{\scriptsize $\pm$12.3} & 58.0{\scriptsize $\pm$8.4} & 56.3{\scriptsize $\pm$2.8} \\
0.8 & 50.2{\scriptsize $\pm$NA} & 50.1{\scriptsize $\pm$NA} & 49.8{\scriptsize $\pm$NA} & 57.3{\scriptsize $\pm$6.4} & 51.5{\scriptsize $\pm$2.3} & 50.2{\scriptsize $\pm$0.9} \\
\rowcolor{Gray} 1.0 & 50.0{\scriptsize $\pm$NA} & 49.7{\scriptsize $\pm$NA} & 50.0{\scriptsize $\pm$NA} & 50.0{\scriptsize $\pm$0.0} & 50.0{\scriptsize $\pm$0.0} & 50.0{\scriptsize $\pm$0.0} \\
1.2 & 50.4{\scriptsize $\pm$NA} & 48.7{\scriptsize $\pm$NA} & 50.8{\scriptsize $\pm$NA} & 49.3{\scriptsize $\pm$3.6} & 50.2{\scriptsize $\pm$0.5} & 56.2{\scriptsize $\pm$1.1} \\
1.4 & 53.6{\scriptsize $\pm$NA} & 47.9{\scriptsize $\pm$NA} & 54.8{\scriptsize $\pm$NA} & 51.0{\scriptsize $\pm$6.8} & 51.1{\scriptsize $\pm$1.2} & 65.2{\scriptsize $\pm$2.5} \\
1.6 & 55.5{\scriptsize $\pm$NA} & 46.7{\scriptsize $\pm$NA} & 62.1{\scriptsize $\pm$NA} & 55.2{\scriptsize $\pm$10.1} & 52.0{\scriptsize $\pm$2.1} & 75.1{\scriptsize $\pm$3.0} \\
1.8 & 60.3{\scriptsize $\pm$NA} & 45.4{\scriptsize $\pm$NA} & 74.0{\scriptsize $\pm$NA} & 61.5{\scriptsize $\pm$13.5} & 53.0{\scriptsize $\pm$3.1} & 83.2{\scriptsize $\pm$4.2} \\
2.0 & 69.9{\scriptsize $\pm$NA} & 43.9{\scriptsize $\pm$NA} & 83.2{\scriptsize $\pm$NA} & 67.2{\scriptsize $\pm$15.9} & 55.0{\scriptsize $\pm$4.6} & 89.3{\scriptsize $\pm$4.2} \\
2.5 & 94.4{\scriptsize $\pm$NA} & 40.1{\scriptsize $\pm$NA} & 92.3{\scriptsize $\pm$NA} & 76.8{\scriptsize $\pm$16.0} & 61.0{\scriptsize $\pm$9.3} & 96.3{\scriptsize $\pm$2.0} \\
3.0 & 98.4{\scriptsize $\pm$NA} & 37.4{\scriptsize $\pm$NA} & 92.7{\scriptsize $\pm$NA} & 83.4{\scriptsize $\pm$13.2} & 64.8{\scriptsize $\pm$11.7} & 98.5{\scriptsize $\pm$0.7} \\
3.5 & 99.3{\scriptsize $\pm$NA} & 35.6{\scriptsize $\pm$NA} & 94.8{\scriptsize $\pm$NA} & 88.5{\scriptsize $\pm$9.7} & 68.5{\scriptsize $\pm$13.2} & 99.1{\scriptsize $\pm$0.4} \\
4.0 & 99.8{\scriptsize $\pm$NA} & 34.3{\scriptsize $\pm$NA} & 97.2{\scriptsize $\pm$NA} & 90.8{\scriptsize $\pm$6.6} & 71.3{\scriptsize $\pm$13.3} & 99.4{\scriptsize $\pm$0.3} \\
4.5 & 100.0{\scriptsize $\pm$NA} & 33.4{\scriptsize $\pm$NA} & 98.0{\scriptsize $\pm$NA} & 91.5{\scriptsize $\pm$4.6} & 73.8{\scriptsize $\pm$12.3} & 99.6{\scriptsize $\pm$0.3} \\
5.0 & 100.0{\scriptsize $\pm$NA} & 32.4{\scriptsize $\pm$NA} & 98.6{\scriptsize $\pm$NA} & 91.4{\scriptsize $\pm$4.2} & 77.1{\scriptsize $\pm$10.7} & 99.6{\scriptsize $\pm$0.3} \\
5.5 & 100.0{\scriptsize $\pm$NA} & 32.0{\scriptsize $\pm$NA} & 98.9{\scriptsize $\pm$NA} & 90.7{\scriptsize $\pm$4.9} & 80.2{\scriptsize $\pm$9.0} & 99.6{\scriptsize $\pm$0.3} \\
6.0 & 100.0{\scriptsize $\pm$NA} & 31.7{\scriptsize $\pm$NA} & 99.0{\scriptsize $\pm$NA} & 89.7{\scriptsize $\pm$5.8} & 82.6{\scriptsize $\pm$8.3} & 99.6{\scriptsize $\pm$0.4} \\
6.5 & 100.0{\scriptsize $\pm$NA} & 31.5{\scriptsize $\pm$NA} & 99.2{\scriptsize $\pm$NA} & 88.6{\scriptsize $\pm$6.7} & 84.3{\scriptsize $\pm$7.9} & 99.6{\scriptsize $\pm$0.5} \\
\midrule
Average &76.8	&41.3	&77.3	&73.0	&67.5	&\textbf{83.7} \\
 \bottomrule
 \end{tabular}}
\vspace{1mm}
\caption{NAS detection performance in regression task (age prediction) for NAS shift of brightness in RSNA boneage dataset measured by AUPR. Highlighted row presents the performance on in-distribution dataset. DE and TAP-MB denotes Deep Ensemble and TAP-Mahalanobis, respectively. $\text{SWAG}^{*}$ = SWAG + Deep Ensemble.  
}
\label{tab:regression_nas_aupr}
\end{table*}
        
\begin{table*}[hbt]
\footnotesize
\centering
\resizebox{0.78\linewidth}{!}{
\begin{tabular}{c  c  c  c c c  c }
\toprule
\multicolumn{1}{c}{\multirow{1}{*}{Brightness}} 
& \multicolumn{3}{c}{\multirow{1}{*}{Baselines}}  & \multicolumn{3}{c}{\multirow{1}{*}{Ours (Task-Agnostic)}} \\  \cmidrule(lr){2-4}  \cmidrule(lr){5-7} 
\multicolumn{1}{c}{} & \multicolumn{1}{c}{DE~\cite{lakshminarayanan2017simple} } & \multicolumn{1}{c}{MC Dropout~\cite{gal2016dropout} } & \multicolumn{1}{c}{$\text{SWAG}^{*}$~\cite{maddox_2019_simple}}  & \multicolumn{1}{c}{TAP-MOS} & \multicolumn{1}{c}{TAP-MB} & \multicolumn{1}{c}{TAPUDD}\\
& & & & (K = 8) & (K = 8) & (Average) 

\\ \midrule
0.0 & 0.0{\scriptsize $\pm$NA} & 100.0{\scriptsize $\pm$NA} & 0.0{\scriptsize $\pm$NA} & 80.0{\scriptsize $\pm$42.2} & 0.0{\scriptsize $\pm$0.0} & 0.0{\scriptsize $\pm$0.0} \\
0.2 & 91.9{\scriptsize $\pm$NA} & 99.2{\scriptsize $\pm$NA} & 94.0{\scriptsize $\pm$NA} & 74.9{\scriptsize $\pm$21.6} & 51.2{\scriptsize $\pm$27.4} & 58.9{\scriptsize $\pm$18.7} \\
0.4 & 94.6{\scriptsize $\pm$NA} & 96.1{\scriptsize $\pm$NA} & 94.7{\scriptsize $\pm$NA} & 71.4{\scriptsize $\pm$23.4} & 69.1{\scriptsize $\pm$18.4} & 90.5{\scriptsize $\pm$2.5} \\
0.6 & 94.7{\scriptsize $\pm$NA} & 95.6{\scriptsize $\pm$NA} & 95.0{\scriptsize $\pm$NA} & 84.0{\scriptsize $\pm$9.6} & 88.9{\scriptsize $\pm$5.4} & 94.4{\scriptsize $\pm$1.0} \\
0.8 & 95.0{\scriptsize $\pm$NA} & 95.6{\scriptsize $\pm$NA} & 95.1{\scriptsize $\pm$NA} & 91.8{\scriptsize $\pm$2.8} & 94.1{\scriptsize $\pm$1.5} & 95.4{\scriptsize $\pm$0.4} \\
\rowcolor{Gray} 1.0 & 95.0{\scriptsize $\pm$NA} & 94.5{\scriptsize $\pm$NA} & 95.0{\scriptsize $\pm$NA} & 95.0{\scriptsize $\pm$0.0} & 95.0{\scriptsize $\pm$0.0} & 95.0{\scriptsize $\pm$0.0} \\
1.2 & 94.7{\scriptsize $\pm$NA} & 95.5{\scriptsize $\pm$NA} & 94.1{\scriptsize $\pm$NA} & 95.2{\scriptsize $\pm$1.2} & 94.5{\scriptsize $\pm$1.4} & 92.6{\scriptsize $\pm$2.1} \\
1.4 & 89.2{\scriptsize $\pm$NA} & 95.6{\scriptsize $\pm$NA} & 93.7{\scriptsize $\pm$NA} & 92.3{\scriptsize $\pm$5.6} & 93.8{\scriptsize $\pm$2.2} & 86.0{\scriptsize $\pm$5.8} \\
1.6 & 78.8{\scriptsize $\pm$NA} & 97.5{\scriptsize $\pm$NA} & 88.6{\scriptsize $\pm$NA} & 87.4{\scriptsize $\pm$9.8} & 92.9{\scriptsize $\pm$3.1} & 74.9{\scriptsize $\pm$9.7} \\
1.8 & 69.7{\scriptsize $\pm$NA} & 98.7{\scriptsize $\pm$NA} & 87.5{\scriptsize $\pm$NA} & 79.8{\scriptsize $\pm$16.8} & 90.3{\scriptsize $\pm$5.3} & 62.1{\scriptsize $\pm$14} \\
2.0 & 53.3{\scriptsize $\pm$NA} & 99.1{\scriptsize $\pm$NA} & 81.7{\scriptsize $\pm$NA} & 73.5{\scriptsize $\pm$24.1} & 88.2{\scriptsize $\pm$7.0} & 49.2{\scriptsize $\pm$18.8} \\
2.5 & 14.9{\scriptsize $\pm$NA} & 100{\scriptsize $\pm$NA} & 60.3{\scriptsize $\pm$NA} & 63.2{\scriptsize $\pm$31.1} & 81.4{\scriptsize $\pm$12.6} & 23.6{\scriptsize $\pm$16.1} \\
3.0 & 6.9{\scriptsize $\pm$NA} & 100{\scriptsize $\pm$NA} & 53.6{\scriptsize $\pm$NA} & 54.5{\scriptsize $\pm$29.9} & 76.7{\scriptsize $\pm$15.4} & 8.8{\scriptsize $\pm$6.8} \\
3.5 & 2.8{\scriptsize $\pm$NA} & 100.0{\scriptsize $\pm$NA} & 35.2{\scriptsize $\pm$NA} & 43.5{\scriptsize $\pm$25.6} & 73.0{\scriptsize $\pm$17.9} & 3.9{\scriptsize $\pm$3.0} \\
4.0 & 0.8{\scriptsize $\pm$NA} & 100.0{\scriptsize $\pm$NA} & 20.2{\scriptsize $\pm$NA} & 36.6{\scriptsize $\pm$20.2} & 69.6{\scriptsize $\pm$19.0} & 2.4{\scriptsize $\pm$2.1} \\
4.5 & 0.0{\scriptsize $\pm$NA} & 100.0{\scriptsize $\pm$NA} & 13.4{\scriptsize $\pm$NA} & 33.7{\scriptsize $\pm$17.2} & 67.3{\scriptsize $\pm$17.6} & 1.3{\scriptsize $\pm$1.6} \\
5.0 & 0.0{\scriptsize $\pm$NA} & 100.0{\scriptsize $\pm$NA} & 8.2{\scriptsize $\pm$NA} & 32.8{\scriptsize $\pm$17.5} & 65.4{\scriptsize $\pm$16.2} & 1.2{\scriptsize $\pm$1.9} \\
5.5 & 0.1{\scriptsize $\pm$NA} & 100.0{\scriptsize $\pm$NA} & 5.7{\scriptsize $\pm$NA} & 34.6{\scriptsize $\pm$20.2} & 61.6{\scriptsize $\pm$16.0} & 1.6{\scriptsize $\pm$3.3} \\
6.0 & 0.0{\scriptsize $\pm$NA} & 100.0{\scriptsize $\pm$NA} & 4.5{\scriptsize $\pm$NA} & 37.7{\scriptsize $\pm$23.0} & 57.7{\scriptsize $\pm$16.6} & 2.1{\scriptsize $\pm$4.5} \\
6.5 & 0.0{\scriptsize $\pm$NA} & 100.0{\scriptsize $\pm$NA} & 3.1{\scriptsize $\pm$NA} & 41.8{\scriptsize $\pm$25.9} & 54.1{\scriptsize $\pm$17.6} & 2.5{\scriptsize $\pm$5.5} \\
\midrule
Average & 44.1	& 98.4	& 56.2	& 65.2	& 73.2	& \textbf{42.3} \\
 \bottomrule
 \end{tabular}}
\vspace{1mm}
\caption{NAS detection performance in regression task (age prediction) for NAS shift of brightness in RSNA boneage dataset measured by FPR95. Highlighted row presents the performance on in-distribution dataset. DE and TAP-MB denotes Deep Ensemble and TAP-Mahalanobis, respectively. $\text{SWAG}^{*}$ = SWAG + Deep Ensemble.   
}
\label{tab:regression_fpr95}
\end{table*}

\end{document}